\documentclass{article}

    \usepackage[preprint]{neurips_2025}

\usepackage[utf8]{inputenc} 
\usepackage[T1]{fontenc}    
\usepackage{url}            
\usepackage{booktabs}       
\usepackage{amsfonts}       
\usepackage{nicefrac}       
\usepackage{microtype}      
\usepackage{xcolor}         
\usepackage{adjustbox}   

\usepackage{transparent}
\usepackage[table]{xcolor}

\usepackage{epsfig}
\usepackage{graphicx}
\usepackage{bbding} 
\usepackage{pifont} 
\usepackage{wrapfig} 
\usepackage{microtype} 
\usepackage{amsfonts}
\usepackage{soul}
\usepackage{color}
\usepackage{bbm}
\usepackage{amsmath,amssymb}
\usepackage{mathrsfs} 
\usepackage{algorithmic,algorithm}
\usepackage{multirow} 
\usepackage{subfigure} 
\usepackage{bbding} 
\usepackage{pifont} 
\usepackage{microtype} 
\usepackage{amsfonts} 
\usepackage{array,caption}
\usepackage{xcolor,colortbl}
\usepackage{soul}
\usepackage{microtype}
\usepackage[pagebackref,breaklinks,colorlinks,citecolor=blue]{hyperref}

\newcommand{\cmark}{\ding{51}}%
\newcommand{\xmark}{\ding{55}}%

\newcommand{\add}[1]{\textcolor{black}{#1}}

\def\sfod{SFOD}
\def\shortmodelname{WSCo}
\def\wie{WIE} 
\def\sie{SIE} 

\definecolor{cmblu}{RGB}{51,102,240}
\definecolor{cmred}{RGB}{241,22,22}

\newtheorem{theorem}{Theorem} 
 
\newtheorem{proof}{Proof} 
 
\newtheorem{proposition}{Proposition} 
\newtheorem{restheo}{Restatement of Theorem} 
\newtheorem{respro}{Restatement of Proposition}

\title{Source-Free Domain Adaptive Object Detection with Semantics Compensation}

\author{%
  Song Tang$^{1,2}$, Jiuzheng Yang$^{1}$, Mao Ye$^{3}$, Boyu Wang$^{4,5}$, Yan Gan$^{6}$, Xiantian Zhu$^{7}$\thanks{Corresponding author: Mao Ye (cvlab.uestc@gmail.com) and Xiatian Zhu (xiatian.zhu@surrey.ac.uk).} \\
  \textsuperscript{1}University of Shanghai for Science and Technology, \textsuperscript{2}Universität Hamburg, \\
  \textsuperscript{3}University of Electronic Science and Technology of China,\\ 
  \textsuperscript{4}Western University, \textsuperscript{5}Vector Institute, \textsuperscript{6}Chongqing University, \textsuperscript{7}University of Surrey\\
}

\begin{document}

\maketitle

\begin{abstract}
Strong data augmentation is a fundamental component of state-of-the-art mean teacher-based Source-Free domain adaptive Object Detection ({\sfod}) methods, enabling consistency-based self-supervised optimization along weak augmentation. However, our theoretical analysis and empirical observations reveal a critical limitation: strong augmentation can inadvertently erase class-relevant components, leading to {\em artificial inter-category confusion}.
To address this issue, we introduce \textbf{W}eak-to-strong \textbf{S}emantics \textbf{Co}mpensation (\textbf{{\shortmodelname}}), a novel remedy that leverages weakly augmented images, which preserve full semantics, as anchors to enrich the feature space of their strongly augmented counterparts. 
Essentially, this compensates for the class-relevant semantics that may be lost during strong augmentation on the fly. 
Notably, {\shortmodelname} can be implemented as a generic plug-in, easily integrable with any existing SFOD pipelines.
Extensive experiments validate the negative impact of strong augmentation on detection performance, and
the effectiveness of {\shortmodelname} in enhancing the performance of previous detection models on standard benchmarks. Our code and data are available at \url{https://github.com/tntek/source-free-domain-adaptive-object-detection}.
\end{abstract}

\section{Introduction}

{Source-Free domain adaptive Object Detection} ({\sfod})~\cite{SED} aims to adapt detection models pre-trained on a source domain to an unlabeled target domain without access to the source training data. 
Current state-of-the-art SFOD methods~\cite{balanced} are based on the Mean Teacher (MT) framework~\cite{meanteacher}, which leverages self-supervised learning through a weak-to-strong data augmentation mechanism. 
In this design, an input image is projected into two data flows using weak and strong augmentations, which are then mapped into their feature space, respectively. 
The resulting region-aligned instance feature pairs, guided by the teacher model's proposals, enable consistency-based self-supervised learning within each pair.



\begin{wrapfigure}{r}{0.55\textwidth}
\centering
\includegraphics[width=0.52\textwidth,trim=0pt 0pt 0pt 0pt, clip]{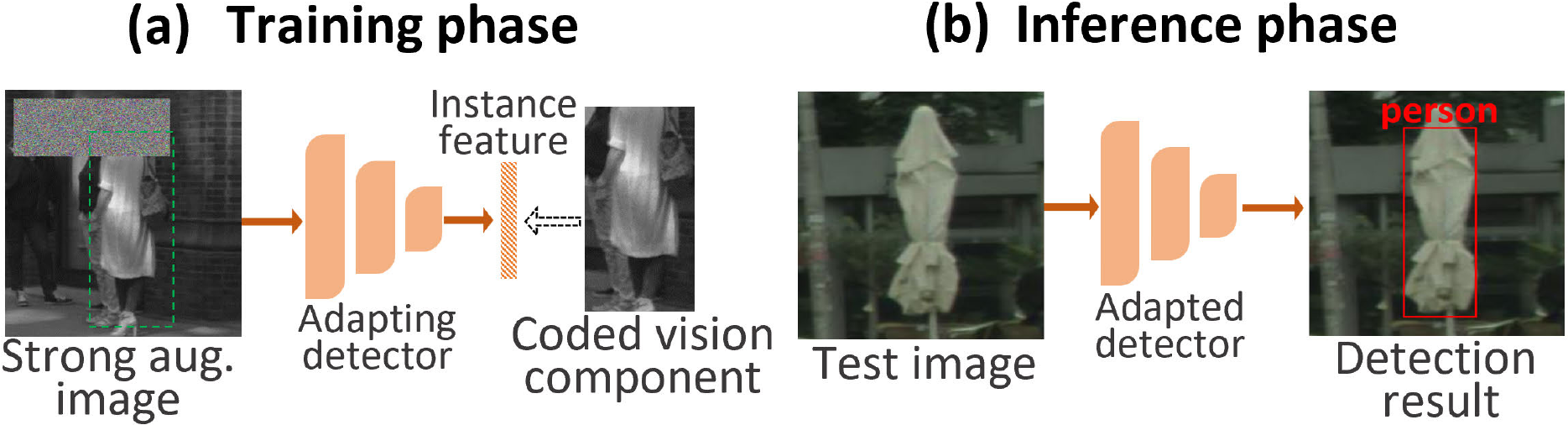}
\caption{Illustration of {artificial inter-category confusion} 
caused by strong data augmentation. 
(a) The head is mosaicked during augmentation.
(b) As a result, the model confuses the classes ``person'' and ``umbrella''.
} 
\label{fig:idea}
\end{wrapfigure}

In this process, strong data augmentation plays a crucial role by creating rich contrasts that promote domain-invariant feature extraction. However, random strong disturbances, such as mosaics, color jittering, and blurring, can erase class-crucial visual components, leading to {\em artificial inter-category confusion}. 
For instance, as shown in Fig.~\ref{fig:idea}, the head is a key discriminative feature to identify the class ``person''. 
If the head is mosaicked, the model may misclassify the image as the class ``umbrella''. 

To further investigate this issue, we analyze the strong augmentation process from an information-theoretic perspective. We find that strong augmentation introduces additional information entropy, which theoretically underlies the inter-class confusion. 
To address this challenge, we propose the \textbf{W}eak-to-strong \textbf{S}emantics \textbf{Co}mpensation (\textbf{{\shortmodelname}}) approach. 
{\shortmodelname} leverages weakly augmented samples, which preserve full semantic information, as references to enhance the representations of their strongly augmented counterparts, thereby recovering visual information lost during strong augmentation. 
Specifically, we construct \add{a semantics shared space} over region-aligned instance feature pairs, forming a Weak Instance Embedding set ({\wie}) and a Strong Instance Embedding set ({\sie}) for weak and strong augmentations, respectively. 
\add{This latent embedding space is learned by a mapping network that is regulated through an adversarial semantics calibration.
We achieve this calibration by applying a gradient approximation incorporating contrastive regularization.}
To enable knowledge transfer from {\wie} to {\sie}, we develop a dynamic pseudo-labeling strategy. This involves establishing and progressively updating a set of prototypes on {\wie} throughout the training process. The pseudo labels from {\wie} are then transferred to {\sie} via region-pair associations, enabling a supervised contrastive learning to enhance the representation of {\sie}.  
\add{Accounting for the object detection task, our contrastive scheme jointly exploits instance and image uncertainty, integrating semantics-rich positive contrast while adaptively removing the negative impact of background. 
}

Our \textbf{contributions} include: 
\textbf{(1)} 
\add{We identify a fundamental issue of artificial inter-category confusion caused by strong data augmentation in state-of-the-art SFOD methods. 
To address this, we develop a theoretical framework for this problem from an information-theoretic perspective.} 
\textbf{(2)} We also introduce a novel mitigating approach, {\shortmodelname}, which recovers class-critical visual information lost during strong augmentation by using weakly augmented samples as alignment references. 
\textbf{(3)} Extensive experiments show that when used as a generic plug-in module during training, {\shortmodelname} significantly enhances the performance of state-of-the-art SFOD models on standard benchmarks.


\section{Related Work}
\textbf{Unsupervised Domain Adaptive Object Detection (UDA-OD).}
Prior UDA-OD methods roughly follow five strategies. The first is adversarial feature learning \cite{DA-faster,SWDA,UDAOD1,UDAOD2,MEGA}, using gradient reversal layers as in DANN \cite{DANN}. The second involves pseudo-labeling \cite{clipart-water,UDAOD4,UDAOD5}, using high-confidence predictions to train the target domain. The third is image-to-image translation \cite{UDAOD6,UDAOD7,UDAOD8,UDAOD9}, converting images between domains using unpaired translation models. The fourth is domain randomization \cite{UDAOD10,UDAOD9}, generates multiple stylized versions of source data for robust training. The fifth is MT training \cite{UMT,MTOR}, which improves generalization by incrementally training with unlabeled data. Despite these advancements, all methods still rely on access to source domain data.

\textbf{Source-Free Domain Adaptive Object Detection.} 
Most {\sfod} methods follow a self-training MT paradigm, broadly categorized into three approaches. The first approach focuses on obtaining higher quality pseudo-labels by setting thresholds through self-entropy descent \cite{SED} or balancing classes \cite{balanced}. The second approach treats target domain images as separate domains by utilizing different variances or augmentations and extracts domain-invariant features via graph-based alignment \cite{LODS} or adversarial learning \cite{SOAP,A2SFOD}. The third approach enhances object representation by contrastive learning via instance relation graph \cite{IRG} or adjacent proposals \cite{LPU}. 
Although achieving impressive results, these methods above lose focus on the artificial inter-category confusion problem.

\section{Method}
\textbf{SFOD problem statement.} 
Suppose the source domain $\mathcal{D}_s = \{(I_i^s, Y_i^s)\}_{i=1}^{N_s}$ is labeled, where $Y_i^s = \{(b_j^s, c_j^s)\}_{j=1}^{M_s}$ denote the bounding boxes and classes of the objects in the $i$-th source image $I_i^s$, $M_s$ denote the total number of object in $I_i^s$, and $N_s$ stands for the total number of source images. 
The target domain $\mathcal{D}_t = \{x_i\}_{i=1}^{N_t}$ is unlabeled, where $N_t$ is the total number of target images, which obey the same distribution, which is different from that of the source domain. 
Our goal is to transfer the source detection model pre-trained on $\mathcal{D}_s$ to the target domain $\mathcal{D}_t$.
During adaptation, $\mathcal{D}_t$ is available while $\mathcal{D}_s$ cannot be accessed.

\subsection{Formulating Artificial Inter-category Confusion in {\sfod}} \label{sec:csl}

We will first theoretically show how strong augmentation leads to {\em artificial inter-category confusion}. \add{Then, we formulate the {\sfod} problem considering artificial inter-category confusion.}



\textbf{Formulation of strong augmentation.}  
Strong augmentations do not exhibit the typical characteristics of affine transformations, such as translation, scaling, rotation, and shear. 
Instead, these operations would impose changes to image attributes (e.g., via color Jitter and RandomGrayscale) and modifications in content appearance (e.g., GaussianBlur and RandomErasing). 
Computationally, this process can be expressed as point-wise alterations to pixel values. 
Mathematically, this process can be depicted by performing a dot multiplication between an image and a mask matrix corresponding to a specific strong augmentation. 
Aligning with this point of view, we have the following proposition. 

\begin{proposition}
\label{prop:strong-aug}
Let random variables $X$ and $\Omega$ be an image and a masking operator corresponding to a specific strong augmentation, respectively. 
The strongly augmented image can be formulated to random variable $X\odot\Omega$ where $\odot$ means the element-wise multiplication. 
\end{proposition}

\textbf{Theoretical results.}  
As the strong augmentation operations are executed in a random fashion, there exists some kind of uncertainty. 
Thus, we conduct a theoretical analysis from the perspective of the information theory.
In this context, we have the following theorem regarding the connection between the information entropy of the strongly augmented image and the final prediction.
\begin{theorem}
\label{thm:stroaug-extrainfor}
Given the strong augmentation process formulated in Proposition~\ref{prop:strong-aug}, $H(\cdot)$ computes information entropy of the input variable, the strongly augmented input is $X'=X\odot\Omega$, and 
$Y \in \mathcal{C}$ is the corresponding label of objects in $X$. 
Assume the classifier produces a predictive distribution $P(Y|X')$. 
If the augmentation operator $\Omega$ destroys or occludes object's key semantic content, then the model output's entropy increases:
\begin{equation}
\label{eqn:thm}
H(Y|X\odot\Omega) = H(Y|X) + H_{\Omega}(Y),
\end{equation}
where $H_{\Omega}(Y)$ is the entropy increase caused by the strong augmentation.
\end{theorem}
{\bf Theorem}~\ref{thm:stroaug-extrainfor} provides a critical insight: {\em Strong augmentations will introduce extra information entropy on the predictive results}, indicating an increased uncertainty in object recognition, giving rise to artificially leading to inter-category confusion.

For the MT-based detection paradigm, strong augmentation presents a double-edged sword. 
While positively increasing True Positives (TP) by introducing more visual contrastive elements, it also leads to an increase in False Positives (FP), which results in undesirable uncertainty $H_{\Omega}(Y)$ (Empirical evidence supporting this phenomenon can be found in \texttt{Section~\ref{sec:emps}}).
Unlike existing approaches that focus on enhancing the positive effects of strong augmentation, our emphasis is on minimizing its negative impact. 
From an information theory perspective, the {\sfod} problem, which considers artificial inter-category confusion, can be expressed as a constrained optimization issue: 
\begin{equation}
\label{eqn:sfod-aic}
\max  I(Y|X\odot\Omega), ~~~ s.t.~ \min H_{\Omega}(Y), 
\end{equation}
where $I(\cdot,\cdot)$ stands for the mutual information function, $Y$ is a random variable that represents the object category in image $X$ predicted by an MT-based detection model.

\vspace{-10pt}
\add{
\subsection{Optimization}
In the standard MT-based detection context, $\max  I(Y|X\odot\Omega)$ in Eq.~\eqref{eqn:sfod-aic} is optimized by performing a self-supervised learning with objective $\mathcal{L}_{\rm{mt}}$~\cite{IRG,ECCV2024LPLD}. 
Formally, as displayed in Fig.~\ref{fig:fw}, the strong branch outputs predictions $\hat{Y}=\{\hat{b}_i,\hat{c}_i\}_{i=1}^{M}$ for the strongly augmented image $\hat{\boldsymbol{I}}$ where $\hat{b_i}$ and $\hat{c}_i$ are the bounding boxes and category distribution of the $i$-the instance in $\hat{\boldsymbol{I}}$. 
Similarly, based on the weakly augmented image $\bar{\boldsymbol{I}}$, the weak branch produces $M$ predictions $\bar{Y}=\{\bar{b}_i,\bar{c}_i\}_{i=1}^{M}$. 
Let $\bar{Y}^h=\{\bar{b}_i^h,\bar{c}_i^h\}_{i=1}^{M}$ be high confident subset in $\bar{Y}$. 
The standard MT objective $\mathcal{L}_{\rm{mt}}$ will be 
\begin{equation}
\label{eqn:loss-mt-std}
\begin{split}
\min_{{\Theta}_{\rm stg}} {{\cal L}_{{\rm{mt}}}}{\rm{ }} = \overbrace {{{\cal L}_{{\rm{rpn}}}}(\hat I,{{\bar b}^h}) + {{\cal L}_{{\rm{rcnn}}}}(\hat I,{{{\bar{a}} }^h})}^{{{\cal L}_{{\rm{det}}}}(\hat I,\;{{\bar Y}^h})} + \overbrace {{1 \over M}\sum {{D_{{\rm{KL}}}}} ({{\hat c}_i}||{{\bar c}_i})}^{{{\cal L}_{{\rm{con}}}}(\hat I,\bar Y)},
\end{split}
\end{equation}
where ${\Theta}_{\rm stg} = \{{\Theta}_{\rm rpn}^{\rm stg}, {\Theta}_{\rm ext}^{\rm stg}, {\Theta}_{\rm rcnn}^{\rm stg}\}$ are model parameters of RPN network, Feature extraction network and RCNN network in the strong branch, respectively; 
$D_{\rm{KL}}$ is the Kullback–Leibler divergence function; $\bar{\boldsymbol{a}}^h$ is the one-hot version of $\bar{c}^h$. 
In Eq.~\eqref{eqn:loss-mt-std}, $\mathcal{L}_{\rm{con}}$ presents a semantic consistency regularization between the region-aligned instance feature pairs.  
$\mathcal{L}_{\rm{det}}$ derives from the Fater-RCNN paradigm, consisting of location regression term ($\mathcal{L}_{\rm{rpn}}$) and classification term ($\mathcal{L}_{\rm{rcnn}}$). 
To exclude the noise in $\bar{Y}$, only credible predictions $\bar{Y}^h$ are treated as pseudo labels involved in training.
}

\add{
For the constraint in Eq.~\eqref{eqn:sfod-aic}, $\Omega$ stands for a combination of a couple of random operations. 
It is difficult to explicitly express or implicitly estimate its probability distributions.  
We therefore cannot optimize $\min H_{\Omega}(Y)$ by the definition method~\cite{IID,prode} or the variational method~\cite{Theim,variationalIM}. 
To address the problem, we provide a semantics compensation solution: Extracting compensation information from weakly augmented samples to enhance the representation of strongly augmented samples with vital visual information lost. 
This is inspired by an observation that in the MT framework, weak augmentations, such as resizing and flipping, only minimally alter the images. 
We realize this idea by designing a method, {\shortmodelname}, as detailed in the next section. 
}

\section{{\shortmodelname} Design}

\begin{figure*}[t]
    \centering
    \includegraphics[width=0.95\textwidth,trim=0pt 0pt 0pt 0pt, clip]{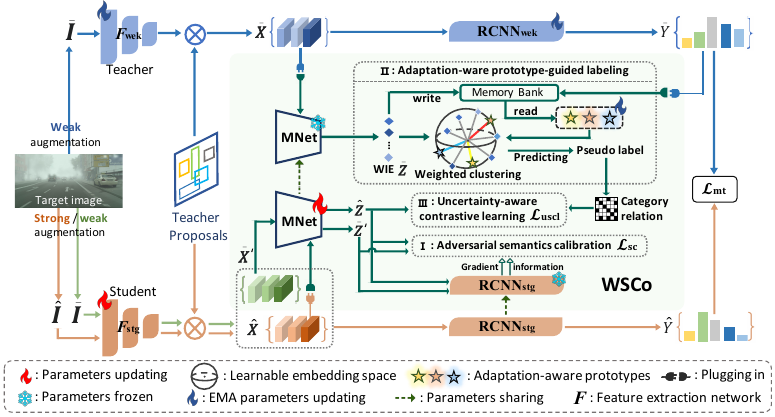} 
    \caption{Overview of {\bf \shortmodelname}. 
    The standard MT framework regularized by objective $\mathcal{L}_{\rm{mt}}$ has generated weak ($\bar{\boldsymbol{X}}$) and strong ($\hat{\boldsymbol{X}}$) instance features.
    In {\shortmodelname} (blocked by green box),   
    a mapping network (termed MNet), which is optimized by {\it I: Adversarial semantics calibration},   
    first projects $\bar{\boldsymbol{X}}$, $\hat{\boldsymbol{X}}$ into a latent space, obtaining semantics bias-less weak and strong instance embedding sets (i.e., {\wie}, {\sie}) $\bar{\boldsymbol{Z}}$, $\hat{\boldsymbol{Z}}$.  
    After that, {\it II: Adaptation-ware prototype-guided labeling} refines semantics in $\bar{\boldsymbol{Z}}$, while {\it III: Uncertainty-ware supervised contrastive learning} enhances the representation of $\hat{\boldsymbol{Z}}$ by encoding the mined semantics.  
    }
    \label{fig:fw}
\end{figure*}

{\bf Model architecture.} 
As illustrated in Fig.~\ref{fig:fw}, {\shortmodelname} is built upon the a standard MT framework with objective $\mathcal{L}_{\rm{mt}}$.
The overall model consists of the weak and strong branches that serve as the teacher and student models, respectively.  
Both branches are based on typical detectors such as Faster-RCNN~\cite{faster}, initialized as the source model, following the existing SFOD methods. 
Specifically, in the strong branch, the strongly augmented image $\hat{\boldsymbol{I}}$ is first converted to corresponding feature maps by the feature extraction network and further to $M$ strong instance features $\hat{\boldsymbol{X}}=\{\hat{\boldsymbol{x}}_i\}_{i=1}^M$ tailored by the teacher proposals (generated by RPN in the weak branch). 
The weak branch is the same as the strong one, except for: (1) inputting the weakly augmented image and (2) Exponential Moving Average (EMA)-wise model updating. 
As the weakly augmented image $\hat{\boldsymbol{I}}$ goes through the weak branch, we also obtain $M$ weak instance features $\bar{\boldsymbol{X}}=\{\bar{\boldsymbol{x}}_i\}_{i=1}^M$. 
\add{Different from the standard MT framework, we also input $\bar{\boldsymbol{I}}$ into the strong branch, generating instance features $\bar{\boldsymbol{X}}'=\{\bar{\boldsymbol{x}}'_i\}_{i=1}^M$.}
Here, by regional correspondence of the teacher proposals, $\bar{\boldsymbol{X}}$ and $\hat{\boldsymbol{X}}$ can form cross-augmentation feature pairs $\langle {\bar{\boldsymbol{X}}, \hat{\boldsymbol{X}}} \rangle =\{(\bar{\boldsymbol{x}}'_i,\hat{\boldsymbol{x}}_i)\}_{i=1}^M$, 
\add{whilst forming pairs $\langle {\bar{\boldsymbol{X}}', \hat{\boldsymbol{X}}} \rangle =\{(\bar{\boldsymbol{x}}_i,\hat{\boldsymbol{x}}_i)\}_{i=1}^M$ over $\bar{\boldsymbol{X}}'$ and $\hat{\boldsymbol{X}}$.}

{\bf In the functional aspect}, {\shortmodelname} serves as a plug-in involving three components. 
\add{Specifically, we first reduce semantic bias between $\langle {\bar{\boldsymbol{X}}, \hat{\boldsymbol{X}}} \rangle$ by projecting them into a semantics shared space. We achieve this by a mapping network (termed MNet) and the \textbf{(I)} adversarial semantic calibration ($\mathcal{L}_{\rm sc}$) building upon $\langle {\bar{\boldsymbol{X}}', \hat{\boldsymbol{X}}} \rangle$. 
The learned MNet projects$\langle {\bar{\boldsymbol{X}}, \hat{\boldsymbol{X}}} \rangle$ into a latent embedding space, creating embedding pairs ${\boldsymbol{Z}}=\langle {\bar{\boldsymbol{Z}}, \hat{\boldsymbol{Z}}} \rangle$, where $\bar{\boldsymbol{Z}}$ and $\hat{\boldsymbol{Z}}$ are {\wie} and {\sie}, respectively.}  
The {\bf(II)} adaptation-aware prototype-guided labeling predicts pseudo-categories for $\bar{\boldsymbol{Z}}$, achieving knowledge refinement from the weak side. 
The {\bf(III)} Uncertainty-aware supervised contrastive learning ($\mathcal{L}_{\rm uscl}$) encodes the mined knowledge into $\hat{\boldsymbol{Z}}$. 
\add{
In this method, we pay attention to high-uncertainty instances, which are discovered by the inconsistency between pseudo labels and neighborhood relations. 
In addition, the image uncertainty is considered to build an adaptive background contrast, filtering out strong noise in background instances. 
}
The details of these three components are presented below.

\subsection{Weak-to-strong Semantics Compensation} 

\textbf{I.~Adversarial semantics calibration.} 
Although generating sharp contrasts, the weak and strong augmentations also inject extra semantics shift/bias into the paired instance features $\langle {\bar{\boldsymbol{X}}, \hat{\boldsymbol{X}}} \rangle$.   
This shift might undermine our cross-augmentation semantics compensation.   
To tackle this issue, we employ MNet to project $\langle {\bar{\boldsymbol{X}}, \hat{\boldsymbol{X}}} \rangle$ into a latent embedding space to reduce the semantics shift between $\bar{\boldsymbol{X}}$ and $\hat{\boldsymbol{X}}$. 
MNet's structure and working details are provided in \texttt{Appendix \ref{app-imp}}.


\add{
Meanwhile, we propose the adversarial semantics calibration method to regulate MNet's parameters. 
Given MNet maps $\langle\bar{\boldsymbol{X}}', \hat{\boldsymbol{X}}\rangle$ to $\boldsymbol{Z}' = \langle\bar{\boldsymbol{Z}}', \hat{\boldsymbol{Z}}\rangle$, the objective is
\vspace{-7pt}
\begin{equation}
\label{eqn:loss-seman-cali}
\footnotesize{
\min_{\Theta_{\rm sc}} \mathcal{L}_{\rm{sc}} = \overbrace{\!1-\!{\rm{cos}}\left(\nabla_{\Theta_{\text{MNet}}} \mathcal{L}_{\rm rcnn}({\bar{Z}}', {\bar a}^h),\nabla_{\Theta_{\text{MNet}}} \mathcal{L}_{\rm rcnn}({\hat{Z}}, {\bar a}^h)\right)}^{\mathcal{L}_{\rm grad}\left({\bar{Z}'},{\hat{Z}}\right)} - \overbrace{\alpha \!\!\sum_{{\boldsymbol{z}}_i \in \boldsymbol{Z}'} \!\log 
\frac{\exp ( {\boldsymbol{z}}_{i} \cdot {\boldsymbol{z}}_{i}^{+} / \tau )} 
{\sum_{{\boldsymbol{z}}_j \in \boldsymbol{Z}',j\neq i} \exp\left({{\boldsymbol{z}}_{i} \cdot {\boldsymbol{z}}_{j} / \tau} \right)}}^{\mathcal{L}_{\rm unsup}\left({\bar{Z}}',{\hat{Z}}\right)},
}
\end{equation}
where
$\Theta_{\rm sc} = \{\Theta_{\rm ext}^{\rm stg},\Theta_{\rm MNet}\}$, $\Theta_{\text{MNet}}$ are MNet's parameters; ${\rm cos}(\cdot,\cdot)$ computes the cosine-similarity; $\nabla_{\Theta_{\text{MNet}}} \mathcal{L}_{\rm rcnn}({\bar{Z}}', {\bar a}^h)$ and $\nabla_{\Theta_{\text{MNet}}} \mathcal{L}_{\rm rcnn}({\hat{Z}}, {\bar a}^h)$ are gradients of classification loss $\mathcal{L}_{\rm rcnn}$ w.r.t $\Theta_{\text{MNet}}$ as inputting ${\bar Z}'$ and ${\hat Z}$, respectively; $\bar{\boldsymbol{a}}^h$ is the high-confidence one-hot coding, same as Eq.~\eqref{eqn:loss-mt-std}, $\tau$ is temperature parameter, positive sample $\boldsymbol{z}_i^{+}$ is identified by the pair relationship indicated in ${\bar Z}'$. 
}

\add{
In Eq.~\eqref{eqn:loss-seman-cali}, $\mathcal{L}_{\rm grad}$ reduces the semantic shift by encouraging a similar optimization direction.
However, under this regulating, the weak and strong features might degenerate to a similar representation. 
We address this problem by introducing the unsupervised contrastive learning regulator $\mathcal{L}_{\rm unsup}$, highlighting the feature discrimination. 
By this adversarial design, we make the successive semantics compensation work upon a semantics-calibrated space.   
}


\add{
\textbf{II. Adaptation-aware prototype-guided labeling.}
Due to the absence of real labels, we utilize weighted clustering to label {\wie} $\bar{\boldsymbol{Z}}$.
In this approach, we employ prototypes, which encode the adaptation dynamics, to guide the weights generation, thereby promoting robust knowledge refinement. In each training iteration, this labeling occurs in two phases as follows. 
}


\textbf{Adaptation-aware prototypes discovery.} 
\add{
We generate the prototypes of foreground categories by employing a memory bank that stores reliable historical embeddings.
Considering ``background" is an artificial conception including multiple categories and often image-specific, we discover the background prototype using online weighted clustering. 
}

\add{
Suppose the memory bank is a queue bundle containing $K$ queues with $D$ size, where $K$ is the object category number of training dataset, and $D$ is the storage length. 
We collectively write them to $\mathcal{M}\!\in\!\mathbb{R}^{K \times D}$. 
For the $k$-th foreground category ($k\!\in\!K$), the prototype in iteration {\it t}, is $\mathcal{P}_k^{t}$, while treating the background as the $(K+1)$-th independent category, its prototype is $\mathcal{P}_{K+1}$. 
Given that the weak branch identifies $M$ instances from the weakly augmented image $\bar{\boldsymbol{I}}$. 
The corresponding instance embeddings and predictions are  $\bar{\boldsymbol{Z}}$ and $\bar{Y}=\{\bar{b}_i,\bar{c}_i\}_{i=1}^{M}$, respectively, where $\bar{b}_i$ and $\bar{c}_i$ are the bounding boxes and category distribution of the $i$-the instance. 
These prototypes can be computed by  
\begin{equation}
\label{eqn:pk-ema}
\mathcal{P}_k^{t} = (1 - \eta) \mathcal{P}_k^{t-1} + \eta \frac{1}{D} \sum_{i=1}^{D} \varTheta_{\text{MNet}}(\mathcal{M}_{k,i}),~~
\mathcal{P}_{K+1} = \frac{\sum_{\bar{\boldsymbol{z}}_i  \in \bar{\boldsymbol{Z}}}  \delta_{K+1}(\bar{c}_i){\bar{\boldsymbol{z}}_{i}}} {\sum_{\bar{\boldsymbol{z}}_i \in \bar{\boldsymbol{Z}}} \delta_{K+1}(\bar{c}_i)},
\end{equation}
where $\eta<1$, $\mathcal{M}_{k,i}$ is the $i$-th element in the $k$-th row $\mathcal{M}_{k}$; softmax function $\delta_{K+1}(\cdot)$ return the $(K+1)$-th element of the output vector. 
}
 
As depicted in Eq.~\eqref{eqn:pk-ema}, $\mathcal{P}_k^{t}$ is updated in the EMA fashion, thereby the dynamics of adaptation are captured. 
Regarding updating the memory bank, for each category, we use the Top-D weak instance embeddings with high confidence to update  $\mathcal{M}$, according to the weak branch's predictions $\bar{Y}^{h}$.


\add{
\textbf{Pseudo-category prediction.} 
Given that the prototypes provide a robust classification basis, labeling $\bar{\boldsymbol{Z}}$ is achieved using similarity comparison in two steps.   
First, we obtain prototypes-based pseudo label $\bar{p}_i$ for $\bar{\boldsymbol{z}}_{i} \in \bar{\boldsymbol{Z}}$ by the nearest centroid measurement: $\bar{p}_i=\arg\min_{j} D_{cos}(\bar{\boldsymbol{z}}_{i}, \mathcal{P}_j)$, where $D_{cos}$ computes the cosine distance. 
Second, the same as~\cite{shot,TANG2022467}, we obtain final pseudo-category using weighted K-Means method where $\bar{\boldsymbol{z}}_{i}$'s weight is $\bar{p}_i$. 
}

\add{
\textbf{III.~Uncertainty-aware contrastive learning.}
Conventional supervised contrastive learning treats all samples from the same category equally, ignoring the different amounts of semantic information that each sample can carry.
This shortcoming is amplified further by the obtained noisy pseudo-categories, due to wrong positive-negative partitions.  
Moreover, for object detection tasks, foreground and background are imbalanced significantly~\cite{oksuz2020imbalance}. 
This combines with the artificial inter-category confusion, introducing considerable semantic noise.
In this paper, we jointly exploit the instance and image uncertainty to mitigate the two problems above.
}


\add{
\textbf{Leveraging instance uncertainty at positive samples level.}
Anti-commonsense samples often provide more information in a learning process. 
Generally, the samples in the same neighborhood geometry could share the same category.  
If a sample exists outside this neighborhood but still belongs to that category, it can be classified as abnormal data with high uncertainty, termed a ``hard positive''. 
Inspired by this concept, we identify hard positives to create semantics-rich positive contrasts.  
}

\add{
For any $\hat{\boldsymbol{z}}_i \in \hat{\boldsymbol{Z}}$, its category is the pseudo-category of the paired $\bar{\boldsymbol{z}}_i \in \bar{\boldsymbol{Z}}$.
Its positive data group $\mathcal{Z}_i \subset \hat{\boldsymbol{Z}}$ shares the same category with $\hat{\boldsymbol{z}}_i$.  
Suppose $\mathcal{Z}_i={\mathcal{Z}_{i}^{hp}} \cup {\mathcal{Z}_{i}^{ep}}$ can be divided into a hard positive set ${\mathcal{Z}_{i}^{hp}}$ and an easy positive set ${\mathcal{Z}_{i}^{ep}}$. 
We obtain ${\mathcal{Z}_{i}^{hp}}$ in two steps: (1) selecting TOP-K ($K=|\mathcal{Z}_i|$) nearest data of $\hat{\boldsymbol{z}}_i$ to construct neighborhood set $\mathcal{N}_{i}$, with the cosine similarity metric; and (2) identifying ${\mathcal{Z}_{i}^{hp}}=\mathcal{Z}_i \setminus (\mathcal{N}_{i} \cap \mathcal{Z}_i)$ (An intuitive illustration is provided in \texttt{Appendix \ref{app-discu}}). 
}

\add{
\textbf{Leveraging image uncertainty at negative samples level.} 
Our design aims at creating negative data set with adaptive background contrasts.
Firstly, we propose an proposal-based image uncertainty estimation method.  
This scheme comes from an intuitive observation. When the detector is unsure about an image, it suggests many possible objects throughout the image. However, when it is more confident, the suggestions focus on specific objects. 
(More discussion is elaborated in \texttt{Appendix \ref{app-discu}})
}


\add{
In practice, our estimation scheme applies Non-Maximum Suppression (NMS) on the teacher proposals for H times where IoU threshold is set to $\varphi=0.1 \sim 0.\rm{H}$, respectively. 
The estimation is achieved by calculating the variance of the retained bounding box numbers, dented by $N_1\sim N_H$, as: $\sigma = \text{Var}(\bar{\boldsymbol{I}}) = \frac{1}{H}\sum_{i=1}^{H}({N_i}-\mu)^2$, 
where $\mu$ is the mean value of $N_1\sim N_H$. 
Of note, the proposal number is sensitive to NMS’s threshold, so we vary the threshold to remove the threshold's bias. 
}

\add{
Subsequently, we obtain the negative data set $\mathcal{B}_i$ for $\hat{\boldsymbol{z}}_i \in \hat{\boldsymbol{Z}}$ by adaptively removing the background instances.
Specifically, the ones in $\mathcal{Z}_i^C \!=\! \hat{\boldsymbol{Z}} \setminus \mathcal{Z}_i$ are adaptively selected as $\mathcal{B}_i\subseteq \mathcal{Z}_i^C$, by the estimated image uncertainty ($\sigma$). 
This selection rule is: $\mathcal{B}_i = \mathcal{Z}_{i}^C \setminus \mathcal{Z}_i^{bg}$ as  $\sigma > u$ and $\mathcal{B}_i=\mathcal{Z}_{i}^C$ as  $\sigma \le u$, where $u$ is a threshold and set $\mathcal{Z}_i^{bg}$ contains instance embeddings assigned to the background category. 
}

\vspace{-10pt}
\add{
\textbf{Regularization.} 
Based on the designs above at positive and negative levels, we build objective as:
\vspace{-4pt}
\begin{equation}
\label{eqn:loss-cis}
\footnotesize{
\min_{{\Theta}_{\rm uscl}} \mathcal{L}_{\rm{uscl}} \!=\! \sum_{\hat{\boldsymbol{z}}_i \in \hat{\boldsymbol{Z}}} \!\!\left[\frac{-\lambda }{|\mathcal{Z}_i^{hp}|} \!\!\sum_{\hat{\boldsymbol{z}}_j \in \mathcal{Z}_i^{hp}}\!\!\log \frac{\exp ( \hat{\boldsymbol{z}}_i \cdot \hat{\boldsymbol{z}}_j / \tau )} {\sum\limits_{\hat{\boldsymbol{z}}_k \in \mathcal{B}_i} \exp ( \hat{\boldsymbol{z}}_i \cdot\hat{\boldsymbol{z}}_k / \tau )}+\frac{-(1-\lambda)}{|\mathcal{Z}_i^{ep}|} \!\!\!\sum_{\hat{\boldsymbol{z}}_j \in\mathcal{Z}_i^{ep}} \!\!\log \frac{\exp ( \hat{\boldsymbol{z}}_i \cdot \hat{\boldsymbol{z}}_j / \tau )} {\sum\limits_{\hat{\boldsymbol{z}}_k \in \mathcal{B}_i} \exp ( \hat{\boldsymbol{z}}_i \cdot \hat{\boldsymbol{z}}_k / \tau )}\!\!\right],
}
\end{equation}
where ${\Theta}_{\rm uscl} = \{{\Theta}_{\rm ext}^{\rm stg}, {\Theta}_{\rm MNet}\}$, $\alpha$ and $\tau$ are trade-off and temperature parameters. 
In the {\sfod} setting, the size of hard and easy positives are often imbalanced (a piece of statistic analysis is provided in \texttt{Appendix \ref{app-discu}}). 
To mitigate mutual misleading, our design specifies the contrastive learning objective as two components. 
The first term in square brackets incorporate the rich semantics from the hard positives. 
However, pseudo-labels provide noise positive grouping. 
Thus, we add the second term, which adopts reliable easy positives, to correct the misleading of the wrongly classified hard positives. 
At the same time, both terms adopt the adaptive background contrasts when selecting negative samples, excluding the significant semantic noise from the background in difficult images. 
}

\textbf{IV. Objective loss and model training.}
As an independent module, {\shortmodelname} can be seamlessly integrated into the standard MT framework or its variations. 
Combing Eq.~\eqref{eqn:loss-mt-std}, Eq.~\eqref{eqn:loss-seman-cali} and Eq.~\eqref{eqn:loss-cis}, the objective with {\shortmodelname} can be generally formulated as: 
\begin{equation}
\label{eqn:obj-final}
\mathcal{L}_{\rm{MT+{\shortmodelname}}} = \mathcal{L}_{\rm{mt}} + \mathcal{R} + \mathcal{L}_{\rm{\shortmodelname}},~ \mathcal{L}_{\rm{\shortmodelname}} = \mathcal{L}_{\rm{sc}} + \beta \mathcal{L}_{\rm{uscl}},
\end{equation}
where
$\alpha$ is a trade-off parameter;
$\mathcal{R}$ is a certain regulating design specified by the MT variations.
The concrete training is summarized to Alg.~\ref{alg:swcl} in \texttt{Appendix~\ref{app-alg}}.

\section{Experiments}
\label{sec:exps}

\textbf{Datasets.}
Our experiments involve seven datasets: \textbf{Cityscapes} \cite{cityscapes}, \textbf{Foggy-Cityscapes} \cite{foggy} (we use only the most severe foggy condition 0.02.), \textbf{ Pascal} \cite{pascal}, \textbf{ Clipart} \cite{clipart-water}, \textbf{ Watercolor} \cite{clipart-water}, \textbf{KITTI} \cite{KITTY} and \textbf{ Sim10K} \cite{sim10k}.  
The details are provided in \texttt{Appendix~\ref{app-dataset}}.
These datasets form five tasks in two scenarios:  
\textbf{(1)} {\it City Scene Adaptation}: Cityscapes $\rightarrow$ FoggyCityscapes, Sim10k $\rightarrow$ Cityscapes, and KITTI $\rightarrow$ Cityscapes; \textbf{(2)} {\it Image Style Adaptation}: Pascal $\rightarrow$ Watercolor and Pascal $\rightarrow$ Clipart.

\textbf{Implementation details.} 
For fair comparison, we follow the experimental settings as~\cite{LODS,IRG,ECCV2024LPLD}. More details are in \texttt{Appendix~\ref{app-imp}}.

\subsection{Empirical Verification of Artificial Inter-Category Confusion} \label{sec:emps}
For an empirical observation, we demonstrate the impact of strong augmentation on the detection process based on variation of two key quantitative indicators: True Positives (TP) and False Positives (FP). 
This demonstration involves three models: Source, SMT and SMT w/ weak-only. 
Among them, Source is pre-trained on the source domain following Fast-RCNN. 
SMT follows the standard MT-based detection pipeline~\cite{IRG,ECCV2024LPLD} with the vanilla detection objective formulated in Eq.~\eqref{eqn:loss-mt-std}. 
SMT w/ weak-only is a variation of SMT where we input the weakly augmented image into both weak and strong branches.

\begin{wrapfigure}{r}{0.55\textwidth}
    \setlength{\belowcaptionskip}{0pt}
    \begin{center}
        \subfigure{\includegraphics[width=0.95\linewidth]{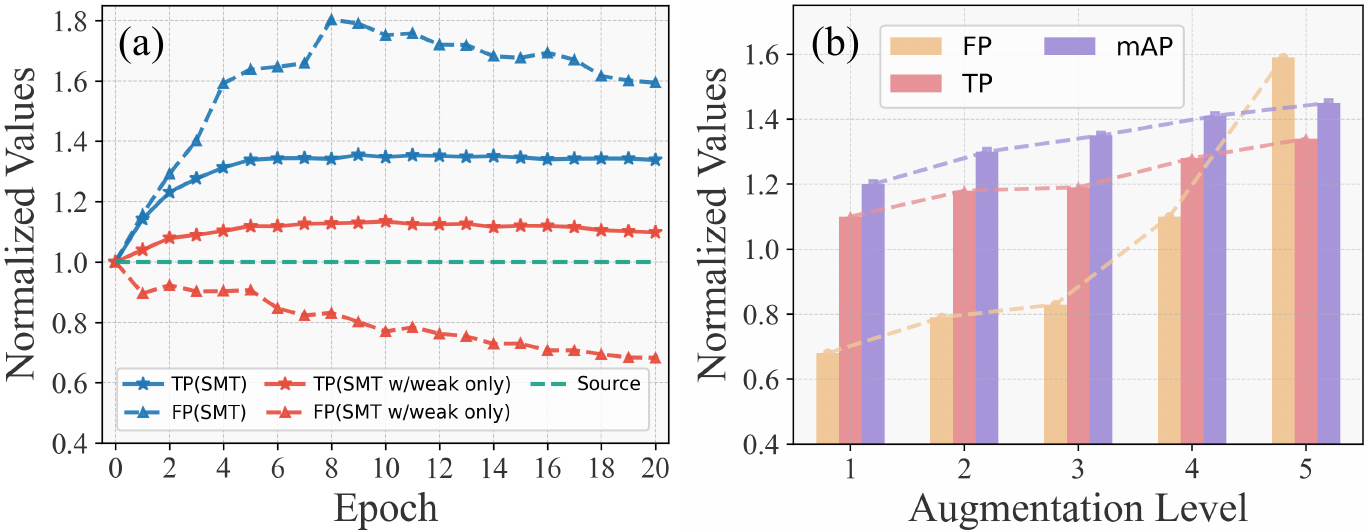}}
    \end{center}
    \vspace{-0.8em}
    \caption{
    TP, FP-based empirical evidence for that strong augmentation causes artificial inter-category confusion.  
    (a) displays results of SMT w/ weak-only (red) and SMT (blue).  
    (b) shows SMT cases with cumulative augmentations. 
    By incremental intensity, the augmentations are five levels (1$\sim$5): 
    Horizontal Flip, ColorJitter, RandomGrayscale, GaussianBlur and RandomErasing.  
    }
    \label{fig:emp-evi}
\end{wrapfigure} 

Here, we display the epoch-wise number variation of TP and FP during SMT w/ weak-only and SMT training, respectively. 
For a clear view, all values are normalized by dividing the corresponding values of Source.
All results are based on task {Cityscapes $\rightarrow$ FoggyCityscapes}.

As shown in Fig.~\ref{fig:emp-evi} (a), the unique weak augmentation in SMT w/ weak-only imposes a positive impact: depress FP while promoting TP (red lines). 
Due to lacking of rich contrast, the promotion in TP is limited. 
The strong augmentation in SMT has a {\it double-edged effect}: It promotes both FP and TP (blue lines). 
The results indicate that strong augmentation will negatively increase FP, which intuitively reflects the artificial inter-category confusion problem. 
   


To figure out the connection between FP and strong augmentation, we also present the cases as cumulatively integrating five kinds of augmentations according to their intensity. 
The proportional varying trend of FP, TP and mAP (mean Average Precision) shown in Fig.~\ref{fig:emp-evi} (b), not only confirms the connection further but also shows that {the strong augmentation essentially promotes trading FP for TP}, which offers a new dimension to understand/evaluate the working mechanism behind the weak-to-strong augmentation strategy.

\begin{table}[t]
    \begin{minipage}[t]{0.53\textwidth}
    \caption{Results on transfer task {Cityscapes $\rightarrow$ FoggyCityscapes}. {SF}: source-free. ResNet-50 is used as the backbone.}
    \centering
    \scriptsize
    \renewcommand{\arraystretch}{0.9} 
    \setlength{\tabcolsep}{0.42mm}{
    \begin{tabular}{llcccccccc|l}
        \toprule
        Methods  & SF  & Pson & Rder & Car & Tuck & Bus & Tain & Mcle & Bcle & mAP \\
        \midrule
        Source  & -- & 31.1 & 38.5 & 36.1 & 19.8 & 23.5 & 9.1 & 21.8 & 30.5 & 26.3 \\
        \midrule
        SWDA \cite{SWDA}   & \xmark & 29.9 & 42.3 & 43.5 & 24.5 & 36.2 & 32.6 & 30.0 & 34.8 & 34.3 \\
        MTOR \cite{MTOR}  & \xmark & 30.6 & 41.4 & 44.0 & 21.9 & 38.6 & 28.0 & 23.5 & 35.6 & 35.1 \\
        SCDA \cite{SCDA}  & \xmark & 33.8 & 42.1 & 52.1 & 26.8 & 42.5 & 26.5 & 29.2 & 34.5 & 35.9 \\
        UMT  \cite{UMT}  & \xmark & 33.8 & 47.3 & 49.0 & 28.0 & 48.2 & 42.1 & 33.0 & 37.3 & 40.4 \\
        MeGA \cite{MEGA} & \xmark & 37.7 & 49.0 & 49.4 & 25.4 & 46.9 & 34.5 & 34.5 & 39.0 & 41.8 \\
        \midrule
        SED  \cite{SED}      & \cmark & 33.2 & 40.7 & 44.5 & 25.5 & 39.0 & 22.2 & 28.4 & 34.1 & 33.5 \\
        SOAP \cite{SOAP}     & \cmark & 35.9 & 45.0 & 48.4 & 23.9 & 37.2 & 24.3 & 31.8 & 37.9 & 35.5 \\
        A$^2$SFOD \cite{A2SFOD}  & \cmark & 32.3 & 44.1 & 44.6 & 28.1 & 34.3 & 29.0 & 31.8 & 38.9 & 35.4 \\
        PETS \cite{PETS}      & \cmark & \textbf{42.0}  & 48.7   & \textbf{56.3}  & 19.3   & 39.3   & 5.5   & \textbf{34.2}  & 41.6 & 35.9 \\
        LPU \cite{LPU}       & \cmark & 39.0 & \textbf{50.3} & 55.4 & 24.0 & \textbf{46.0} & 21.2 & 30.3 & \textbf{44.2} & 38.8 \\
        BT  \cite{balanced}       & \cmark & 38.4 & 47.1 & 52.7 & 24.3 & 44.6 & 36.3 & 30.2 & 40.1 & 39.5 \\
        \midrule
        SMT  \cite{meanteacher}      & \cmark & 34.0 & 43.0 & 45.0 & 23.7 & 25.1 & 25.1 & 31.5 & 32.6 & 36.3 \\
        {\textbf{SMT+\shortmodelname}}    & \cmark & 36.9 & 47.2 & 52.0 & \textbf{30.9} & 45.4 & 39.9 & 29.8 & 42.3 & 40.6 {\tiny\color{blue} (\textuparrow 4.3)} \\ 
        LODS  \cite{LODS}     & \cmark & 34.0 & 45.7 & 48.2 & 27.3 & 39.7 & 19.6 & 32.3 & 37.8 & 37.8 \\
        {\textbf{LODS+\shortmodelname}}  & \cmark & 37.0 & 47.1 & 51.3 & 27.2 & 41.4 & 36.9 & 32.2 & 39.1 & 39.0 {\tiny\color{blue} (\textuparrow 1.2)} \\ 
        IRG   \cite{IRG}    & \cmark & 37.4 & 45.2 & 51.9 & 24.4 & 39.6 & 25.2 & 31.5 & 41.6 & 37.1 \\
        {\textbf{IRG+\shortmodelname}}   & \cmark & 37.6 & 43.2 & 52.0 & 27.1 & 43.2 & 36.2 & 33.6 & 39.8 & 39.1 {\tiny\color{blue} (\textuparrow 2.0)} \\
        LPLD  \cite{ECCV2024LPLD}   & \cmark & 36.4 & 47.1 & 52.2 & 27.3 & 45.7 & 40.6 & 30.7 & 39.4 & 39.9 \\
        {\textbf{LPLD+\shortmodelname}}  & \cmark & 36.1 & 47.2 &52.3 &31.2& 45.0& \textbf{41.1} & 30.8 & 41.7 & \textbf{40.7} {\tiny\color{blue} (\textuparrow 0.8)}\\
    \bottomrule
    \end{tabular}}
    \label{tab:foggy}
    \end{minipage}
    \hspace{0.02\textwidth}
    \begin{minipage}[t]{0.44\textwidth}
    \caption{Results on transfer task {Sim10k $\rightarrow$ Cityscapes} and {KITTI $\rightarrow$ Cityscapes}. ResNet-50 is used as the backbone.}
    \centering
    \scriptsize
    \renewcommand{\arraystretch}{0.91} 
    \setlength{\tabcolsep}{1.0mm}{
    \begin{tabular}{llll}
        \toprule
         &  & \multicolumn{1}{c}{\texttt{Sim10K $\rightarrow$ City}} & \multicolumn{1}{c}{\texttt{KITTI $\rightarrow$ City}} \\
         \cmidrule(lr){3-3} \cmidrule(lr){4-4}
            Method  & SF & AP on car & AP on car \\
        \midrule
            Source  & -- & 33.3 & 34.6 \\  
        \midrule
            SWDA \cite{SWDA} & \xmark & 40.1 & 37.9 \\ 
            SCDA \cite{SCDA} & \xmark & 43.0 & 42.5 \\ 
            UMT  \cite{UMT} & \xmark & 43.1 & -- \\ 
            MeGA \cite{MEGA} & \xmark & 44.8 & 43.0 \\ 
        \midrule 
            SED   \cite{SED}     & \cmark & 43.1 & 44.6 \\  
            SOAP  \cite{SOAP}     & \cmark & 41.6 & 42.7 \\  
            A$^2$SFOD \cite{A2SFOD} & \cmark & 44.0 & 44.9 \\  
            LPU \cite{LPU}       & \cmark & 48.4 & 47.3 \\  
            BT  \cite{balanced}       & \cmark & 48.6 & 48.7 \\ 
        \midrule 
            SMT \cite{meanteacher}       &\cmark & 43.9 & 45.4 \\  
            \textbf{SMT+\shortmodelname}      &\cmark & 49.5 {\tiny\color{blue} (\textuparrow 5.6)}  & 50.7 {\tiny\color{blue} (\textuparrow 5.3)} \\
            LODS  \cite{LODS}     &\cmark & 46.4 & 46.6 \\ 
            {\textbf{LODS+\shortmodelname}}  &\cmark & 48.9 {\tiny\color{blue} (\textuparrow 2.5)}  & 49.4 {\tiny\color{blue} (\textuparrow 2.8)}\\
            IRG  \cite{IRG}      &\cmark & 45.2 & 46.9 \\ 
            {\textbf{IRG+\shortmodelname}}   &\cmark & 51.4 {\tiny\color{blue} (\textuparrow 6.2) } & 50.3 {\tiny\color{blue} (\textuparrow 3.4)}\\
            LPLD \cite{ECCV2024LPLD}      &\cmark & 49.4 & 51.3 \\ 
            {\textbf{LPLD+\shortmodelname}}  &\cmark &\textbf{51.5} {\tiny\color{blue} (\textuparrow 2.1)} & \textbf{53.8} {\tiny\color{blue} (\textuparrow 2.5)}\\
        \bottomrule
    \end{tabular}}
    \label{tab:city}
    \end{minipage}
\end{table}


\subsection{Comparison with State-of-the-art Methods}

\textbf{Competitors.}
To evaluate the effectiveness of {\shortmodelname}, we select 19 competitors in three categories. 
{\it The first} contains Source model pre-trained on the source domain (lower bound) and 
Oracle model trained on target data by ground truth (upper bound). 
{\it The second} has 10 SFOD methods, including SMT (i.e., standard MT approach)~\cite{meanteacher}, SED \cite{SED}, SOAP \cite{SOAP}, LODS \cite{LODS}, A$^2$SFOD \cite{A2SFOD}, IRG \cite{IRG}, LPU \cite{LPU}, PETS \cite{PETS}, BT \cite{balanced} and LPLD~\cite{ECCV2024LPLD}. 
{\it The last} includes 7 UDAOD methods, including SWDA \cite{SWDA}, MTOR \cite{MTOR}, SCDA \cite{SCDA}, MeGA \cite{MEGA}, SAPNet \cite{sapnet}, PD \cite{PD} and UMT \cite{UMT}. 
We integrate {\shortmodelname} with four existing SFOD models:
SMT+{\shortmodelname}, LODS+{\shortmodelname}, IRG+{\shortmodelname}, LPLD+{\shortmodelname}.

\begin{wraptable}{r}{0.6\textwidth}
    \centering
    \scriptsize
    \renewcommand{\arraystretch}{0.9} 
    \caption{Results on {Pascal $\rightarrow$ Watercolor}, {Pascal $\rightarrow$ Clipart} (P $\rightarrow$ C); the full results on {Clipart} are in \texttt{Appendix~\ref{app-quanti}}.}
    \setlength{\tabcolsep}{0.50mm}
    \begin{tabular}{clccccccc|l|l}
        \toprule
        \multicolumn{3}{c}{} & \multicolumn{7}{c}{\texttt{Pascal $\rightarrow$ Watercolor}} & \multicolumn{1}{c}{\texttt{P$\rightarrow$C}} \\
        \cmidrule(lr){4-10} \cmidrule(lr){11-11}
        \cmidrule(lr){4-10} \cmidrule(lr){11-11}
        {} & Methods  & SF & bike & bird & car & cat & dog & prsn & mAP & mAP\\
        \midrule         
        \multirow{12}{*}{\rotatebox{90}{ResNet-101}} & Source  & - & 70.8 & 46.4 & 47.7 & 30.4 & 29.8 & 53.8 & 48.5 & 30.9 \\ 
        & SWDA \cite{SWDA}     & \xmark & 82.3 & 55.9 & 46.5 & 32.7 & 35.5 & 66.7 & 53.3 & 38.1 \\ 
        & SAPNet \cite{sapnet}   & \xmark & 81.1 & 51.1 & 53.6 & 34.3 & 39.8 & 71.3 & 55.2 & 42.2 \\ 
        & PD \cite{PD}       & \xmark & 95.8 & 54.3 & 48.3 & 42.4 & 35.1 & 65.8 & 56.9 & 42.1 \\ 
        & UMT \cite{UMT}      & \xmark & 88.2 & 55.3 & 51.7 & 39.8 & 43.6 & 69.9 & 58.1 & 44.1 \\ 
        \cmidrule(lr){2-10} \cmidrule(lr){11-11}
        
        & {SMT} \cite{meanteacher}    & \cmark & 83.8 & 54.3 & 53.7 & 31.5 & 34.7 & 64.2 & 53.7 & 35.8 \\ 
        & {\textbf{SMT+\shortmodelname}}   & \cmark & 78.5 & 55.2 & \textbf{56.7} & 43.2 & 38.6 & 75.2 & 57.9 {\color{blue} (\tiny\textuparrow 4.2)} & 41.3 {\tiny\color{blue} (\textuparrow 5.5)} \\
        & {LODS} \cite{LODS}   & \cmark & \textbf{95.2} & 53.1 & 46.9 & 37.2 & \textbf{47.6} & 69.3 & 58.2 & 45.2 \\ 
        & {\textbf{LODS+\shortmodelname}} & \cmark & 85.3 & \textbf{55.3} & 56.3 & \textbf{49.9} & 39.0 & \textbf{75.4} & \textbf{60.2} {\tiny\color{blue} (\textuparrow 2.0)}& \textbf{47.3} {\tiny\color{blue} (\textuparrow 2.1)} \\
        \midrule
        \multirow{7}{*}{\rotatebox{90}{{ResNet-50}}}  & Source  & - & 68.8 & 46.8 & 37.2 & 32.7 & 21.3 & 60.7 & 45.7 & 26.8 \\ 

        & {SMT} \cite{meanteacher}     & \cmark & 65.8 & 51.5 & 47.6 & 37.3 & 38.5 & 69.8 & 52.7 & 31.4 \\ 
        & {\textbf{SMT+\shortmodelname}}   & \cmark & 75.4 & 51.1 & 56.2 & 40.1 & 46.8 & \textbf{72.0} & 56.9 {\tiny\color{blue} (\textuparrow 4.2)} & 35.6 {\tiny\color{blue} (\textuparrow 4.2)} \\ 
        & {IRG}  \cite{IRG}   & \cmark &75.9 &52.5 &50.8 &30.8 &38.7 &69.2 &53.0 &31.5 \\ 
        & {\textbf{IRG+\shortmodelname}}  & \cmark & \textbf{76.4} & 51.1 & 55.6 & 41.5 & 39.6 & 69.6 & 55.6 {\tiny\color{blue} (\textuparrow 2.6)} &  33.6 {\color{blue} (\tiny\textuparrow 2.1)} \\
        & {LPLD} \cite{ECCV2024LPLD}   & \cmark & 69.2 & \textbf{53.4} & 54.6 & 37.9 & 45.7 & 70.5 & 55.2 & 34.0 \\ 
        \bottomrule
    \end{tabular}
    \label{tab:water and clipar}
\end{wraptable}

\textbf{Main results.}
The results on the two scenarios are presented in Tab.~\ref{tab:foggy}$\sim$\ref{tab:water and clipar}. 
The best results appear in the group of the methods equipped with {\shortmodelname}.  
In the first scenario, LPLD+{\shortmodelname} improved by \textbf{0.8\%}, \textbf{2.1\%}, and \textbf{2.5\%} in tasks Cityscapes $\rightarrow$ FoggyCityscapes, Sim10k $\rightarrow$ Cityscapes, and KITTI $\rightarrow$ Cityscapes, respectively, compared with the previous best SFOD method LPLD. 
In the second scenario, as adopting the backbone of ResNet-101, LODS+{\shortmodelname} outperforms the second best SFOD method LODS by \textbf{2.0\%} and \textbf{2.1\%} on the task Pascal $\rightarrow$ Watercolor and Pascal $\rightarrow$ Clipart, respectively.
When switching to ResNet-50, LPLD+{\shortmodelname} improve \textbf{2.5\%} (on Pascal $\rightarrow$ Watercolor) and \textbf{1.8\%} (on Pascal $\rightarrow$ Clipart), respectively, compared with previous best SFOD alternative LPLD. 
{\bf Importantly}, all methods equipped with {\shortmodelname} significantly promote the original base methods on all transfer tasks of the two scenarios. 
More qualitative results are presented in \texttt{Appendix~\ref{app-vis}}.

We also made an interesting observation that the performance improvement in the SMT group is greater than that in the other groups utilizing additional designs. 
This phenomenon is explainable: the extra designs might diminish the effectiveness of {\shortmodelname}.
Specifically, the generative style augmentation, which serves as a strong augmentation in LODS, is a relatively low-intensive operation compared with the conventional ones, e.g., RandomErasing, breaking {\shortmodelname}'s working condition.  
The contrastive regularization in IRG is confined to the strong instance features, conflicting with the optimization of {\shortmodelname}. 
The cross-augmentation weighted prediction alignment in LPLD takes low-confident instance pairs into account for diversity enhancement. However, it also introduces noise inevitably, hindering our knowledge refinement from the weak side.

\subsection{Explanation for {\shortmodelname}'s Effectiveness}
\textbf{Protocol.} Aligning with the dimension of trading FP for TP initiated in \texttt{\ref{sec:emps}}, we propose a new metric, called {FP-Gain}, to quantify the performance of MT-based approaches. 
Suppose the source model produces $N_{\rm{TP}}^S$ TPs and $N_{\rm{FP}}^S$ FPs; similarly, the in-training target model has $N_{\rm{TP}}^t$ TPs and $N_{\rm{FP}}^t$ FPs at epoch $t$. 
The FP-Gain at $t$, denoted by $G_{\rm{FP}}^t$, can be computed by $G_{\rm{FP}}^t = \Delta_{\rm{TP}} / \Delta_{\rm{FP}} = (N_{\rm{TP}}^t - N_{\rm{TP}}^S)/(N_{\rm{FP}}^t - N_{\rm{FP}}^S)$. 
Essentially, FP-Gain suggests how much TP improvement can be brought by a unit increase of FN.

\begin{figure*}[t]
    \setlength{\belowcaptionskip}{0pt}
    \begin{center}
        \includegraphics[width=0.95\linewidth]{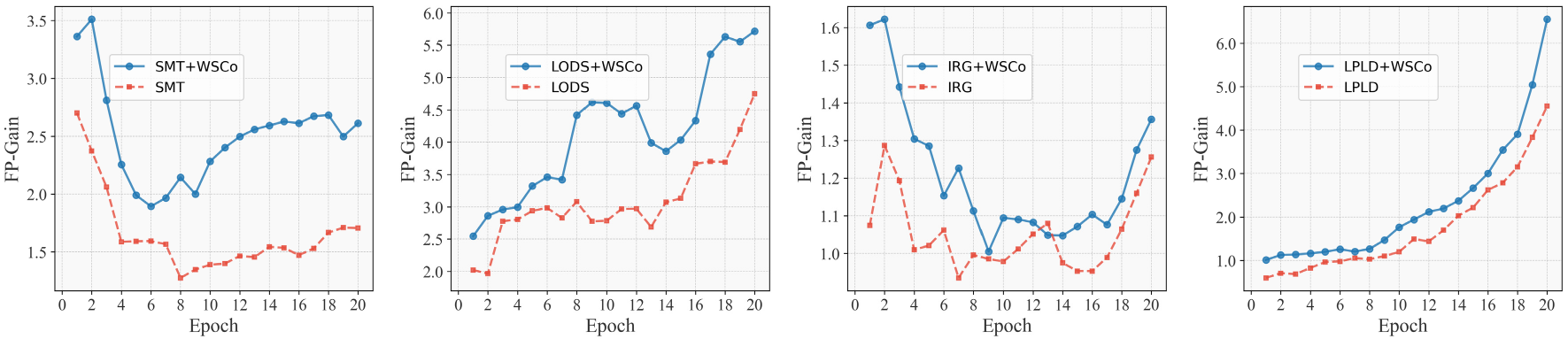}~~~~
    \end{center}
    \caption{
    {FP-Gain analysis on {Cityscapes $\rightarrow$ FoggyCityscapes}. \textbf{Left} to \textbf{Right} display results of SMT, LODS, IRG, and LPLD groups.} 
    }
    \label{fig:fg}
\end{figure*}

\textbf{FP-Gain comparison.} Based on task {Cityscapes $\rightarrow$ FoggyCityscapes}, we present the FP-Gain varying of the four comparison groups. 
As illustrated in Fig.~\ref{fig:fg}, the FP-Gain of the methods with {\shortmodelname} is almost more significant than these base methods over all epochs, besides some individually isolated epochs. 
This indicates that the effect of {\shortmodelname} stems from the increase in FP-Gain, providing a quantitative explanation for why {\shortmodelname} can solve the artificial inter-category confusion problem.  



\subsection{Ablation study} \label{sec:model-anly}
To evaluate the proposed WSCo method, we conduct model analysis based on the SMT group, which includes SMT and SMT+WSCo, without incorporating other specific designs.
All experiments are executed on the tasks {Cityscapes $\rightarrow$ FoggyCityscapes} and {Pascal $\rightarrow$ Clipart}.

This part first isolates the effect of the proposed loss components.
As shown in the top of Tab.~\ref{tab:Ablation_Study} (2--5 row), removing the proposed $\mathcal{L}_{\rm sc}$ or $\mathcal{L}_{\rm uscl}$ leads to a performance decrease compared with the complete version (5 row), confirming the their effect.

\begin{wraptable}{r}{0.47\textwidth}
\centering
\scriptsize
\caption{Ablation study results of mAP on adaptation tasks {Cityscapes $\rightarrow$ FoggyCityscapes} (C $\rightarrow$ F) and {Pascal $\rightarrow$ Clipart} (P $\rightarrow$ C).}
\renewcommand{\arraystretch}{0.9} 
\setlength{\tabcolsep}{1.5mm}{
\begin{tabular}{l|l|ccc|cc}
\toprule 
\# & Methods & $\mathcal{L}_{\rm{mt}}$ & $\mathcal{L}_{\rm{sc}}$ &  $\mathcal{L}_{\rm{uscl}}$ & C$\rightarrow$F & P$\rightarrow$C \\
\midrule
1 &Source & \xmark & \xmark & \xmark  & 26.3 & 30.9 \\
\midrule
2 &SMT \cite{meanteacher}    & \cmark & \xmark & \xmark  & 36.3 & 35.8 \\
\midrule
3 &       & \cmark & \cmark & \xmark  &  39.6   & 38.7    \\ 
4 &       & \cmark & \xmark & \cmark  &  37.8   & 38.1    \\
5 &
\textbf{SMT + {\shortmodelname}}    & \cmark & \cmark & \cmark & \textbf{40.6} & \textbf{41.3} \\
\midrule
6 &\multicolumn{4}{l}{\textbf{SMT+{\shortmodelname} w/o MNet}}    & 39.2  & 38.4 \\
7 &\multicolumn{4}{l}{\textbf{SMT+{\shortmodelname} w/o TwoTerm}} & 39.5  & 39.4 \\
\bottomrule
\end{tabular}}
\label{tab:Ablation_Study}
\end{wraptable}

Subsequently, we evaluate two key designs in {\shortmodelname}: (1) the mapping network, MNet, and (2) the two-term design in $\mathcal{L}_{\rm uscl}$.  
To this end, we create two SMT+{\shortmodelname} variations. 
Among them, SMT+{\shortmodelname} w/o MNet removes MNet, conducting contrastive learning on instance features $\langle {\bar{\boldsymbol{X}}, \hat{\boldsymbol{X}}} \rangle$.  
SMT+{\shortmodelname} w/o TwoTerm adopts the standard supervised contrastive learning object~\cite{khosla2020supervised}, but weighting the hard and easy positives respectively with typical weights, e.g., (0.7, 0.3). 
As shown in Tab.~\ref{tab:Ablation_Study}, SMT+{\shortmodelname} surpasses SMT+{\shortmodelname} w/o MNet by {\bf 1.4}\% on FoggyCityscapes and {\bf 2.9}\% on Clipart. 
The weighted version (7 row) leads to mAP reduction by {\bf 1.1}\% and {\bf 1.9}\% on FoggyCityscapes and Watercolor, respectively, compared to SMT+{\shortmodelname}. 
These results confirm the two designs' effect.

\section{Conclusion} 
In this paper, we explore the problem of artificial inter-category confusion caused by strong augmentation in SFOD approaches.
Firstly, we theoretically demonstrate that strong augmentation leads to this phenomenon by the information theory.  
Following that, to mitigate this issue, we introduce a {\shortmodelname} approach, which enhances the representations of the strong side with crucial visual component loss by integrating compensation information refined from the weak side with full semantics. 
Furthermore, we implement this design to a generic plug-in. 
Extensive experiments show that {\shortmodelname} can effectively promote the performance of the previous SFOD method, offering a general mechanism mitigating artificial inter-category confusion for traditional MT framework. 

\bibliographystyle{plain}
\bibliography{example_paper}

\newpage
\appendix

\section*{Acknowledgments} 
This work is partly funded by the National Natural Science Foundation of China (62476169, 62206168, 62276048); the Postdoctoral Fellowship Program of CPSF, China (GZC20233323).



\section{Proof of Theorem 1} \label{app:proof}

\begin{respro}
\label{prop:strong-aug-1}
Let random variables $X$ and $\Omega$ be an image and a masking operator corresponding to a specific strong augmentation, respectively. 
The strongly augmented image can be formulated to random variable $X\odot\Omega$ where $\odot$ means the element-wise multiplication. 
\end{respro}

\begin{restheo} \label{the:error-effect}  
Given the strong augmentation process formulated in Proposition~\ref{prop:strong-aug}, $H(\cdot)$ computes information entropy of the input variable, the strongly augmented input is $X'=X\odot\Omega$, and 
$Y \in \mathcal{C}$ is the corresponding label of objects in $X$. 
Assume the classifier produces a predictive distribution $P(Y|X')$. 
If the augmentation operator $\Omega$ destroys or occludes object's key semantic content, then the model output's entropy increases:
\begin{equation}
\label{eqn:thm}
H(Y|X\odot\Omega) = H(Y|X) + H_{\Omega}(Y),
\end{equation}
where $H_{\Omega}(Y)$ is the entropy increase caused by the strong augmentation. 
\end{restheo}

\begin{proof} 
If the augmentation $\Omega$ erases semantically meaningful content, then the mutual information between the input and the label decreases $I(X\odot\Omega;Y) < I(X;Y)$.
Applying the mutual information identity $H(Y|X')=H(Y)-I(X';Y)$, we obtain $H(Y|X\odot\Omega) = H(Y) - I(X\odot\Omega;Y)$. 
Since $I(X\odot\Omega;Y) < I(X;Y)$, it follows that:
\begin{equation}
    \label{eqn:final-3}
    \begin{split}
    H(Y|X\odot\Omega) > H(Y|X) = H(Y) - I(X;Y).
    \end{split}
\end{equation}
Turning the above inequality into an equation, we have $H(Y|X\odot\Omega) = H(Y|X) + H_{\Omega}(Y)$, 
where $H_{\Omega}(Y)$ is the entropy increase caused by the strong augmentation.
\end{proof}

\section{Complement Discussion for {\shortmodelname}'s Details} \label{app-discu} 

\begin{figure}[t]
    \begin{minipage}[t]{0.47 \textwidth}
    \centering
    \includegraphics[width=0.98\textwidth,trim=0pt 0pt 0pt 0pt, clip]{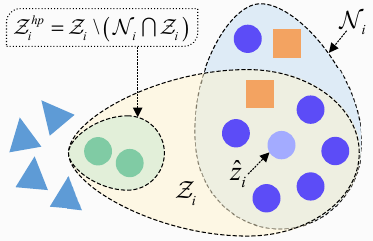} 
    \caption{Illustration of discovering hard positives ${\mathcal{Z}}_i^{hp}$ for $\hat{\boldsymbol{z}}_i$ by inconsistency between $\hat{\boldsymbol{z}}_i$'s neighborhood ${\mathcal{N}}_i$ and positive data set ${\mathcal{Z}}_i$.}
    \label{fig:hp}
    \end{minipage}
    ~~
    \begin{minipage}[t]{0.55\textwidth}
    \centering
    \includegraphics[width=0.98\textwidth,trim=0pt 0pt 0pt 0pt, clip]{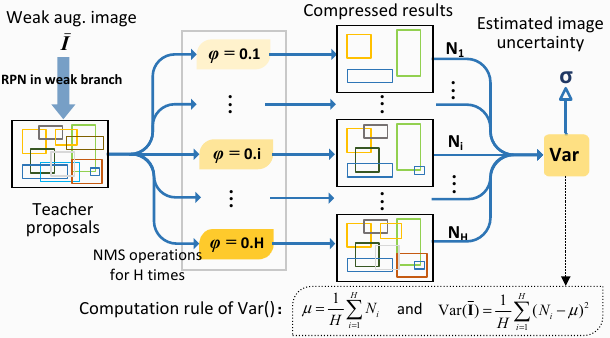}
    \caption{Illustration of proposal-based image uncertainty estimation. $\varphi$ is IoU threshold of Non-Maximum Suppression (NMS).}
    \label{fig:uncertianty}
    \end{minipage}
\end{figure}

\textbf{Hard positives identification.}
For an intuitive view, we demonstrate what the hard positives are and how to discover them in Fig.~\ref{fig:hp}.
Without loss of generality, we take the case of strong embedding $\hat{\boldsymbol{z}}_i \in \hat{\boldsymbol{Z}}$ as an example.  

\textbf{Motivation of proposal-based image uncertainty estimation.} 
With accurate category information, introducing background semantics can boost contrastive learning. 
However, in the scenarios lacking effective supervision, e.g., SFOD, the background information often contains much noise, deteriorating contrastive performance. 
Thus, we design the uncertainty-based adaptive mechanism, as shown in Fig.~\ref{fig:uncertianty}, to balance the background's effect.

\textbf{Statistical analysis of size imbalance between hard and easy positives.} 
To measure the size imbalance between hard and easy positives, we introduce a metric called HE ratio-score. For a given instance \(\hat{\boldsymbol{z}_i} \in \hat{\boldsymbol{Z}}\), the HE ratio-score is defined as \(r_i^{he} = |\mathcal{Z}_i^{hp}| / |\mathcal{Z}_i^{ep}|\), where $\mathcal{Z}_i^{hp}$ and  $\mathcal{Z}_i^{ep}$ are hard and easy positive sets. 
This metric allows us to quantify the size imbalance.   
Concretely, we take a randomly selected image from the FoggyCityscapes dataset and input it into the source detection model to obtain category labels and feature instances. 
We then creat $\mathcal{Z}_i^{hp}$, $\mathcal{Z}_i^{ep}$ by splitting positive set $\mathcal{Z}_i$ according to the method illustrated in Fig.~\ref{fig:hp}.   
After calculating the HE ratio-score for these instances,  we present these results in form of score distributions and category views. 
Additionally, we apply the same statistical analysis to all images in FoggyCityscapes to obtain dataset-level results. 

\begin{figure*}[t]
\centering
\includegraphics[width=1\textwidth]{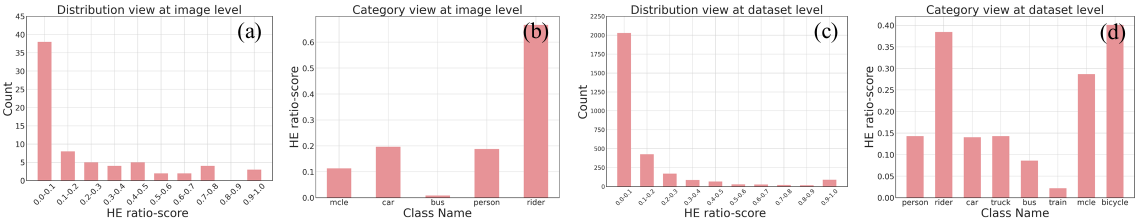} 
\caption{
Statistical analysis of size imbalance between hard and easy positives based on the proposed metric of Hard-Easy (HE) ratio-score. 
(a) and (b) present the results at the image-level, whilst the dataset-level results are given in (c) and (d). 
}
\label{fig:two-terms} 
\end{figure*}

Fig.~\ref{fig:two-terms} shows the statistical analysis results. 
As shown in sub-figures (a) and (c), whether at the image- or dataset-level, the score distributions are biased. 
Specifically, the data within the range of 0.4$\sim$0.5 and 0.5$\sim$0.6 constitute a very small proportion of the total. 
For example, this proportion is {\bf 9.86}\% at the image-level (see (a)), whilst being {\bf 3.21}\% at the data-level (see (c)).   
Additionally, as a validation, the analysis from the category view shows that the HE ratio scores do not approach 0.5, as depicted in sub-figures (b) and (d). 
The results confirm the size imbalance between hard and easy positives, thereby justifying our two-term design in $\mathcal{L}_{\rm uscl}$.


\section{Model Training} \label{app-alg}
Based on the objective in Eq.~\eqref{eqn:obj-final}, we achieve the model training. 
During model training, we optimize the strong branch iteration-wise, while the EMA updating on the weak branch is triggered epoch-wise. 
The overall training process is summarized as Alg.~\ref{alg:swcl}.

\begin{algorithm}[t]
\caption{Training pseudo code of methods equipped with {\shortmodelname}}
\label{alg:swcl}
\begin{algorithmic}[1]
\REQUIRE Unlabeled target dataset $\mathcal{X}_t$, source model $\varTheta_{\text{s}}=\{{\Theta}_{\rm rpn}^{s}, {\Theta}_{\rm ext}^{s}, {\Theta}_{\rm rcnn}^{s}\}$, MNet $\varTheta_{\text{MNet}}$. 
\STATE \textbf{Initialisation:} Set weak (teacher) and strong (student) branches $\varTheta_{\text{wek}} = \varTheta_{\text{stg}}= \varTheta_{\text{s}}$, initialize $\varTheta_{\text{MNet}}$ randomly. 
\FOR{$m=0 \rightarrow EpochNum$}
    \STATE Update $\varTheta_{\text{wa}}$ in a EMA manner; 
    \FOR{$t=0 \rightarrow IterNum$}
    
    \STATE Sample a target image $\boldsymbol{I}$ from $\mathcal{X}_t$;
    
    \STATE Generate weakly and strongly augmented images $\bar{\boldsymbol{I}}$ and $\hat{\boldsymbol{I}}$;  
    
    \STATE Generate instance features $\bar{\boldsymbol{X}}$, $\hat{\boldsymbol{X}}$ and predictions $\hat{Y}$, $\bar{Y}$ by letting $\hat{\boldsymbol{I}}$ and $\bar{\boldsymbol{I}}$ go through $\varTheta_{\text{wek}}$ and $\varTheta_{\text{stg}}$, respectively; 
    
    \STATE Form paired features $\langle {\bar{\boldsymbol{X}}, \hat{\boldsymbol{X}}} \rangle$ by the teacher proposals; 
    

    \STATE Generate paired instance embeddings ${\boldsymbol{Z}}=\langle {\bar{\boldsymbol{Z}}, \hat{\boldsymbol{Z}}} \rangle$ by inputting $\langle {\bar{\boldsymbol{X}}, \hat{\boldsymbol{X}}} \rangle$ into fixed $\varTheta_{\text{MNet}}$; 
    
    \STATE Generate pseudo-categories for $\bar{\boldsymbol{Z}}$ by the adaptation-aware prototype-guided labeling; 
    
    \STATE Estimate proposal-based uncertainty ($\sigma$) of $\boldsymbol{I}$; 
    
    \STATE Propagate pseudo-categories from $\bar{\boldsymbol{Z}}$ to $\hat{\boldsymbol{Z}}$ by the pair relationship that ${\boldsymbol{Z}}$ depicted; 
    
    \STATE Form positive-negative division $(\mathcal{Z}_i, \mathcal{B}_i) \in \hat{\boldsymbol{Z}}$ for any $\hat{\boldsymbol{z}}_i \in \hat{\boldsymbol{Z}}$; 
    
    \STATE Split $\mathcal{Z}_i$ to hard positive set $\mathcal{Z}_i^{hp}$ and easy positive $\mathcal{Z}_i^{ep}$; 

    \STATE Generate embedding $\bar{\boldsymbol{Z}}'$ by letting $\bar{\boldsymbol{I}}$ go through $\varTheta_{\text{ext}}^{\text{stg}}$ and $\varTheta_{\text{MNet}}$;

    \STATE Form paired embedding $\langle {\bar{\boldsymbol{Z}}', \hat{\boldsymbol{Z}}} \rangle$ by the teacher proposals;

    \STATE Update $\varTheta_{\text{stg}}$ and $\varTheta_{\text{MNet}}$ by optimizing objective $L_{\rm{MTG+{\shortmodelname}}}$ over $\langle {\bar{\boldsymbol{Z}}', \hat{\boldsymbol{Z}}} \rangle$. 
    
    \STATE Update $\varTheta_{\text{wek}}$ epoch-wise. 
    \ENDFOR
\ENDFOR\\
{\bf Return} Adapted $\varTheta_{\text{sa}}$ model.
\end{algorithmic}
\end{algorithm}

\section{Datasets}
\label{app-dataset}
We evaluate the proposed method on five adaptation tasks, building on seven datasets listed below. 
    \begin{itemize}

    \item \textbf{Cityscapes} \cite{cityscapes} dataset includes 2,975 training images and 500 test images captured under normal weather conditions, with annotations for 8 categories. 
    \item \textbf{FoggyCityscapes} \cite{foggy} simulates foggy conditions using images from Cityscapes and retains the same annotations. For the adaptation task Cityscapes $\rightarrow$ FoggyCityscapes, we following the common setup \cite{LODS}, only use the most severe foggy condition (0.02) for model training and evaluation.
    
    \item \textbf{Pascal} \cite{pascal} is a dataset of natural images containing 20 categories, we follow the standard data split as described in \cite{SWDA}, selecting the training and validation sets from PASCAL VOC 2007 and 2012 as the source domain, which together include 16,551 images.
    
    \item  \textbf{Clipart} \cite{clipart-water} contains 1,000 clipart-style images across the same 20 categories as Pascal, with 500 images allocated for training and 500 for testing. For the adaptation task Pascal $\rightarrow$ Clipart, we utilizeits training and testing image to train and test our model correspondingly, where the source model was trained on Pascal.
    
    \item  \textbf{Watercolor}\cite{clipart-water} dataset consists of 1K training images and 1K testing images across six categories. For the adaptation task Pascal $\rightarrow$ Watercolor, we follow the common setup \cite{LODS}, training the source model using only the six categories shared between the Pascal and Watercolor datasets. 
    
    \item  \textbf{KITTI} \cite{KITTY} dataset contains 7,481 urban images that differ from typical urban scenes and are used to train the source detector model, with Cityscapes as the target domain. For the adaptation task KITTI $\rightarrow$ Cityscapse, we following the common setup \cite{LODS}, using the model trained on all source data to detect the car category in the target domain (Cityscapes).
    
    \item \textbf{Sim10K} \cite{sim10k} is a synthetic dataset obtained from the video game Grand Theft Auto V (GTA5), containing 10K images of the car category with 58,701 bounding boxes. For the adaptation task Sim10k $\rightarrow$ Cityscapse, we report the performance on the car category as a common setting  in the target domain (Cityscapes).
\end{itemize}

\section{Implementation Details}
\label{app-imp}

\textbf{Model integration.} 
For a fair comparison, we integrate {\shortmodelname} with the selected base methods by specifying the regularization $\mathcal{R}$ in $\mathcal{L}_{\rm{MT+{\shortmodelname}}}$ (Eq.~\eqref{eqn:obj-final}.  
Specifically, $\mathcal{R}$ in IRG is the contrastive regularization based on strong instance features' similarity, whilst $\mathcal{R}$ in LPLD is the weighted prediction aligning regularization crossing the weak and strong sides. 
As for LODS, $\mathcal{R}$ is a feature-level aligning, along with the generative style augmentation.

\textbf{Training setting.}
For the sake of fairness, we follow the experimental setting of previous work~\cite{LODS,IRG}, where Faster RCNN is adopted as the base detector. The backbone network is ResNet \cite{resnet} pre-trained on ImageNet \cite{imagenet}. In all experiments, the shorter side of each input image is resized to 600 pixels. For the proposed framework, the EMA momentum rate for the teacher network is set to $0.9$ and the numbers of teacher proposals is set to $M = 300$. Additionally, the high-confidence threshold generated by the teacher network is set to 0.9. The student model is trained using SGD optimizer with the learning rate of 0.001 and the momentum of 0.9. We report the mAP metric of the teacher network on the target domain with an IoU threshold of 0.5 during test. All experiments were implemented on a single 4090 GPU using the PyTorch and Detectron2 detection frameworks, with a batch size of 1 and trained for 10 epochs.

\textbf{Network setting.}
For the \textit{Urban Scene Adaptation} scenario (Cityscapes $\rightarrow$ FoggyCityscapes, Sim10k $\rightarrow$ Cityscapes, and KITTI $\rightarrow$ Cityscapes), we use ResNet50 as backbone. 
As for the \textit{Image Style Adaptation} scenario (Pascal $\rightarrow$ Watercolor and Pascal $\rightarrow$ Clipart), we use ResNet50 and ResNet101, aligning with the comparison methods for a fair evaluation.

\textbf{Augmentations.}
Following previous works \cite{IRG, ECCV2024LPLD}, weak augmentations include Resizing and HorizontalFlip, while strong augmentations include ColorJitter, RandomGrayscale, GaussianBlur, and RandomErasing. 
Specifically, for LODS+{\shortmodelname}, we leverage style enhancement as strong augmentations, the same as LODS \cite{LODS}.

\textbf{Parameter setting.}
For all transfer tasks, we adopt the same parameter settings. 
The trading off parameters $\alpha$ (in Eq.~\eqref{eqn:loss-seman-cali}), $\lambda$ (in Eq.~\eqref{eqn:loss-cis}), and $\beta$ (in Eq.~\eqref{eqn:obj-final}) are set to 0.1, 0.5, and 0.5, respectively. 
In addition, the uncertainty estimation threshold $u$ is set to 20.0, the temperature parameter for contrastive learning $\tau$ is set to 0.07. 
As for the memory bank, the length size and EMA weight are set to $D = 10$ and $\eta = 0.4$, respectively (see Eq.~\eqref{eqn:pk-ema}).

\begin{wraptable}{r}{0.45\textwidth}
\caption{Architectures of the MNet.}
\renewcommand{\arraystretch}{1.3} 
\centering
\scriptsize
\setlength{\tabcolsep}{3mm}{
\begin{tabular}{c}
\hline
{\bf Structure of Mapping Network} \\
\hline
Conv 2048 $\times$ 3 $\times$ 3, stride 2 $\rightarrow$ BatchNorm $\rightarrow$ ReLU \\
Conv 1024 $\times$ 3 $\times$ 3, stride 2 $\rightarrow$ BatchNorm $\rightarrow$ ReLU \\
Conv 1024 $\times$ 3 $\times$ 3, stride 2 $\rightarrow$ BatchNorm $\rightarrow$ ReLU \\
FC (1024 , 2048) $\rightarrow$ BatchNorm $\rightarrow$ ReLU\\
FC (2048 , 2048)\\
FC (2048 , 512)\\
\hline
\end{tabular}
}
\label{tab:mpnet}
\end{wraptable}

\textbf{Structure and working details of MNet.}
The architecture details of the MNet is presented in Tab.~\ref{tab:mpnet}. 
In terms of MNet's working, there are three details. 
The first one is MNet serves for both branches in different ways. 
Specifically, MNet is jointly trained with the strong branch, while it is frozen as working for the weak branch. 
This alternative way encourages a gradual search for the optimal shared space. 
The second one is that MNet only works during the training phase while being removed as the inference time. 
The last one is the warm-up skill employed in the first training epoch, which reduces the impact of random initialization on clustering. 
Specifically, the method of adaptation-aware prototype-guided clustering is not performed on the weak instance embeddings at the beginning of the training but serves in a gradual transition manner as follows.

\begin{figure*}[t]
\centering
\includegraphics[width=0.24\textwidth]{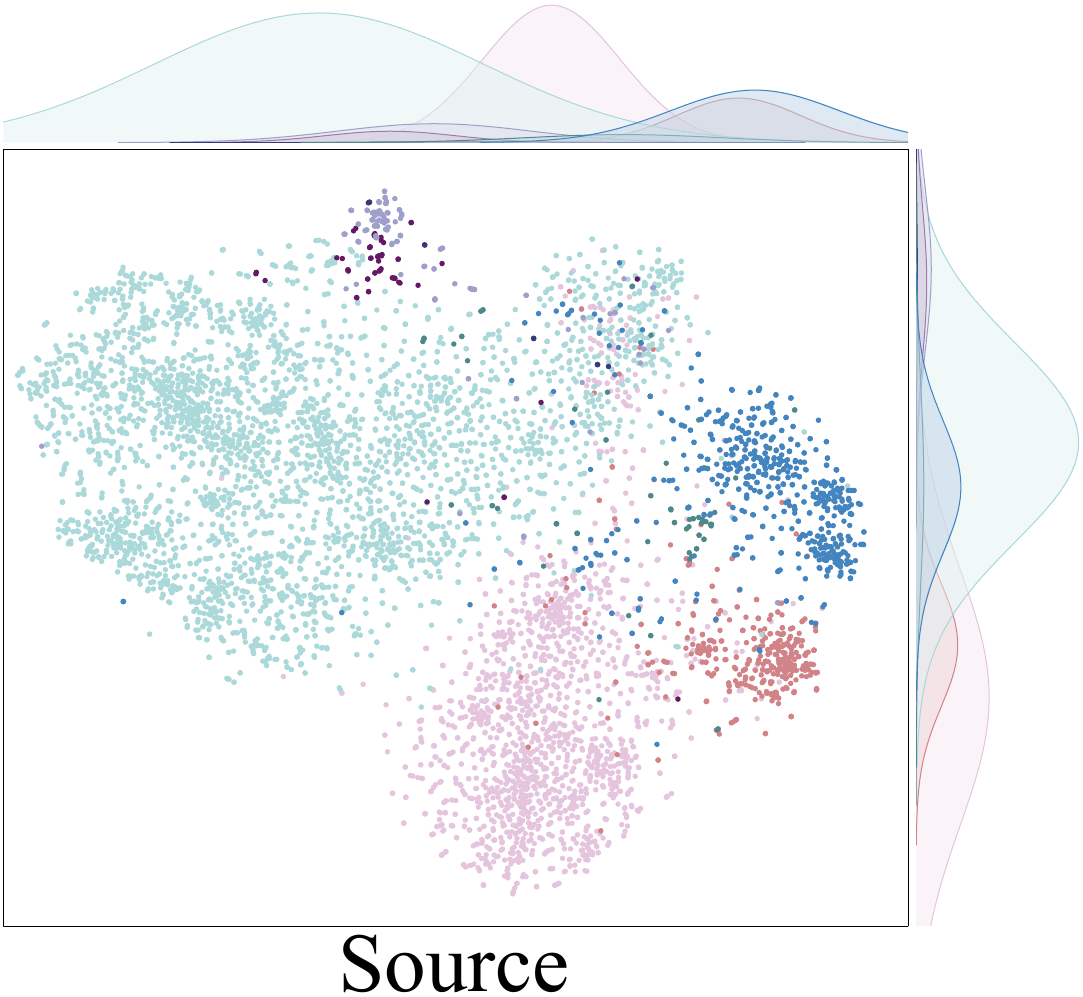} 
\includegraphics[width=0.24\textwidth]{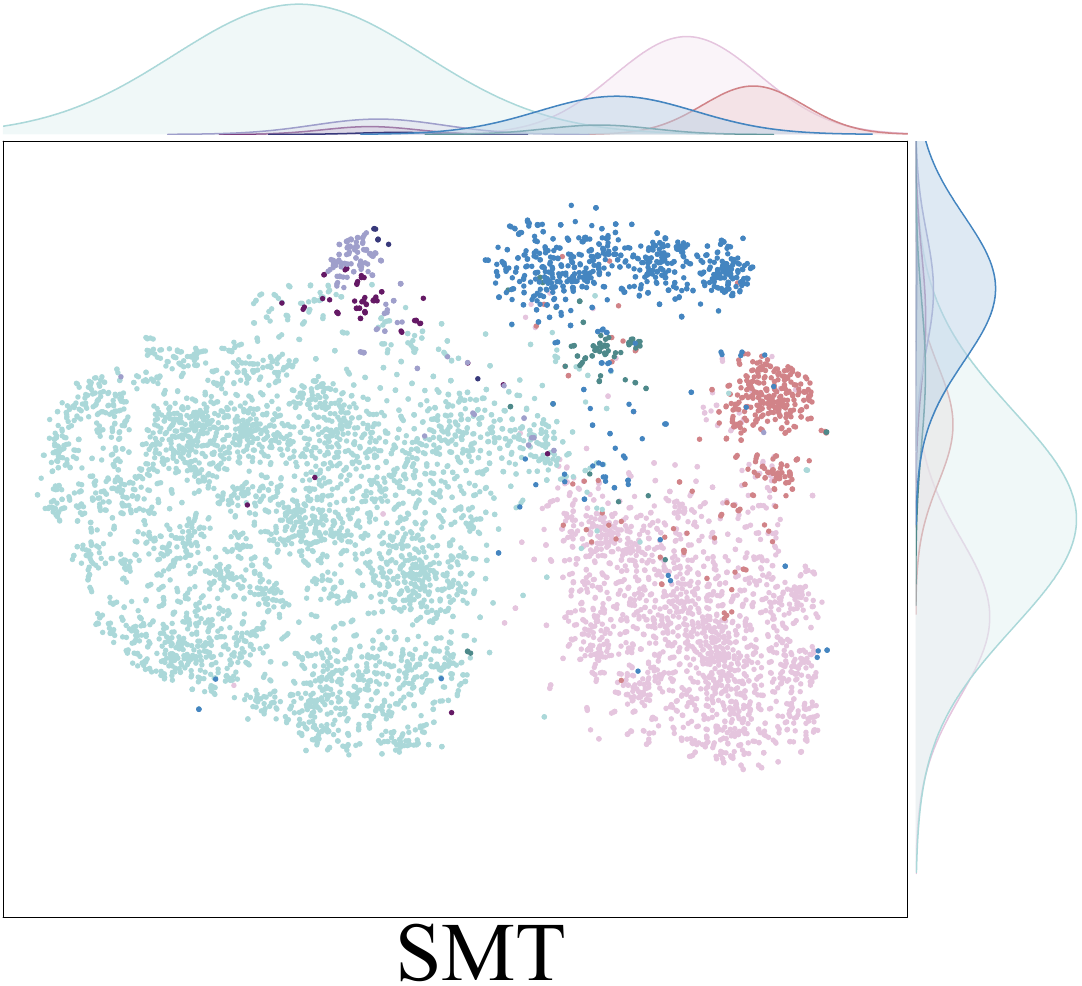} 
\includegraphics[width=0.24\textwidth]{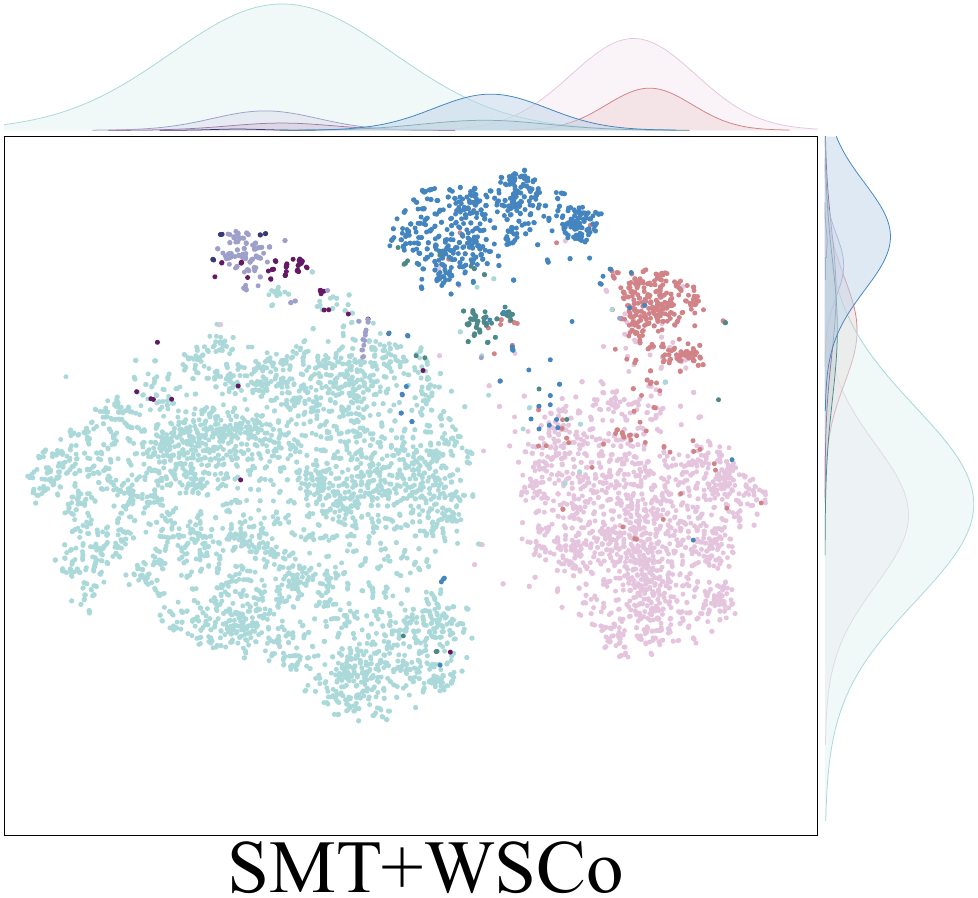} 
\raisebox{-0.02cm}{\includegraphics[width=0.24\textwidth]{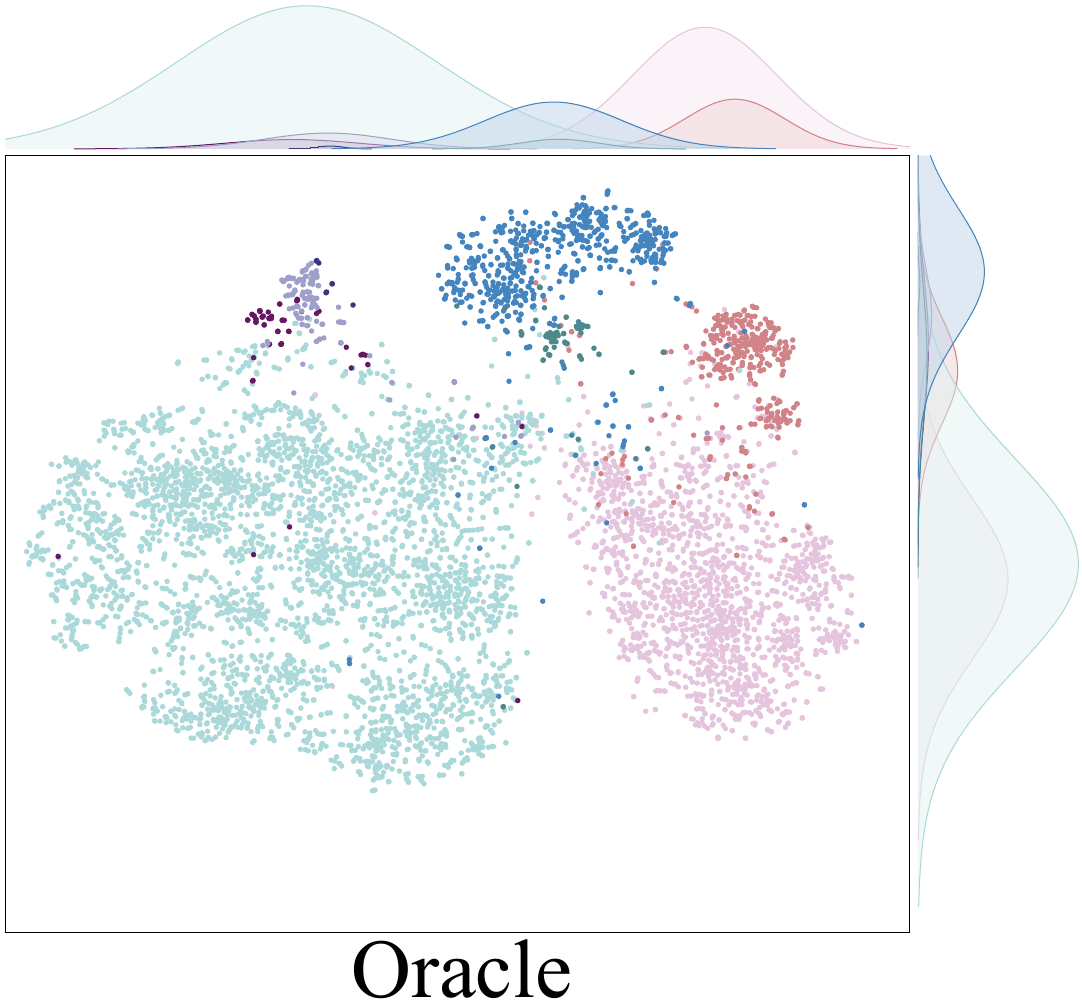}}
\caption{Feature distribution visualization on task {Cityscapes $\rightarrow$ FoggyCityscapes} by the t-SNE tool. 
Categories are presented in different colors.}
\label{fig:tsne-dis}
\end{figure*}

\begin{figure*}[t]
\centering
\includegraphics[width=0.92\textwidth]{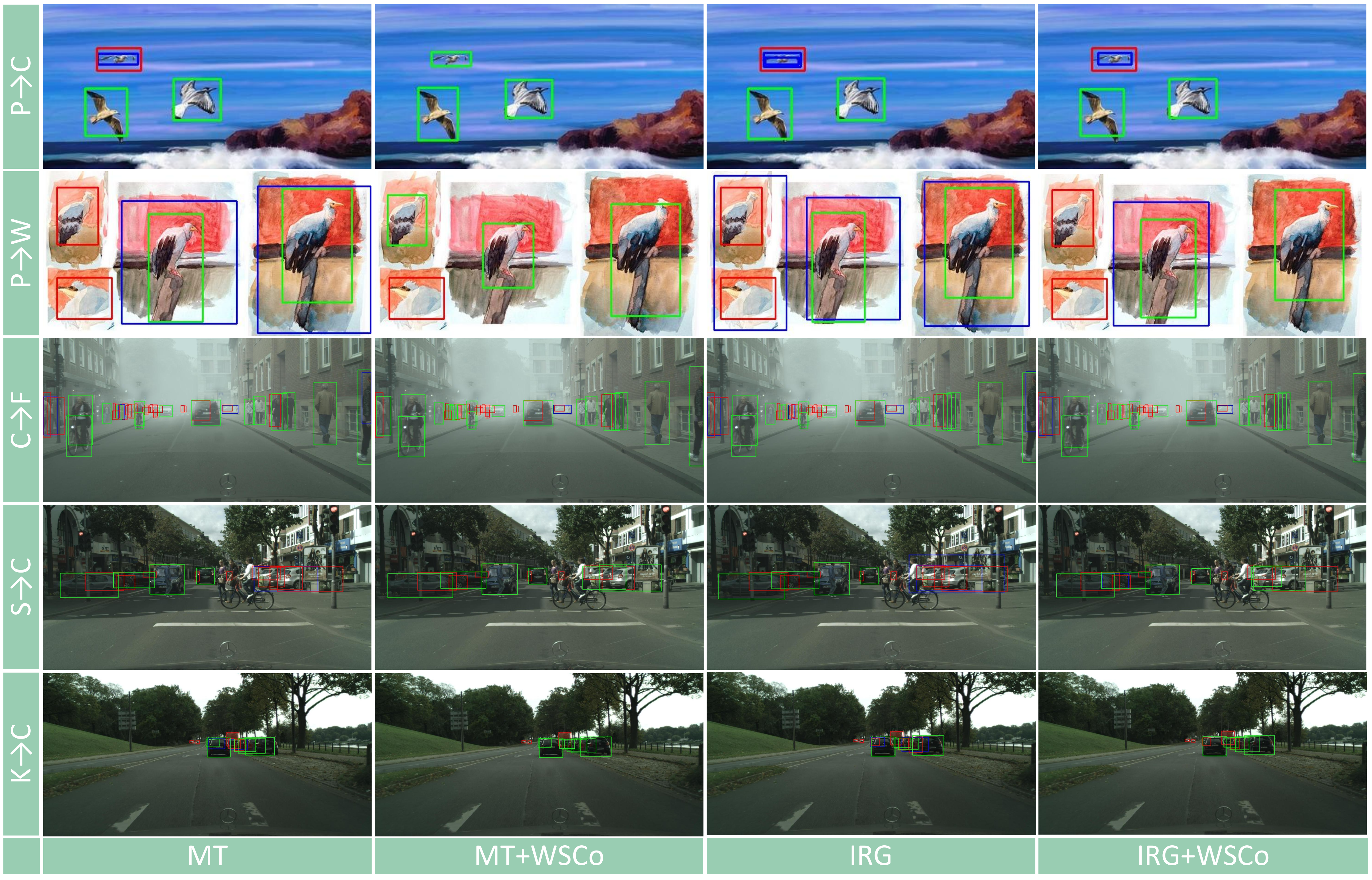} 
\includegraphics[width=0.92\textwidth]{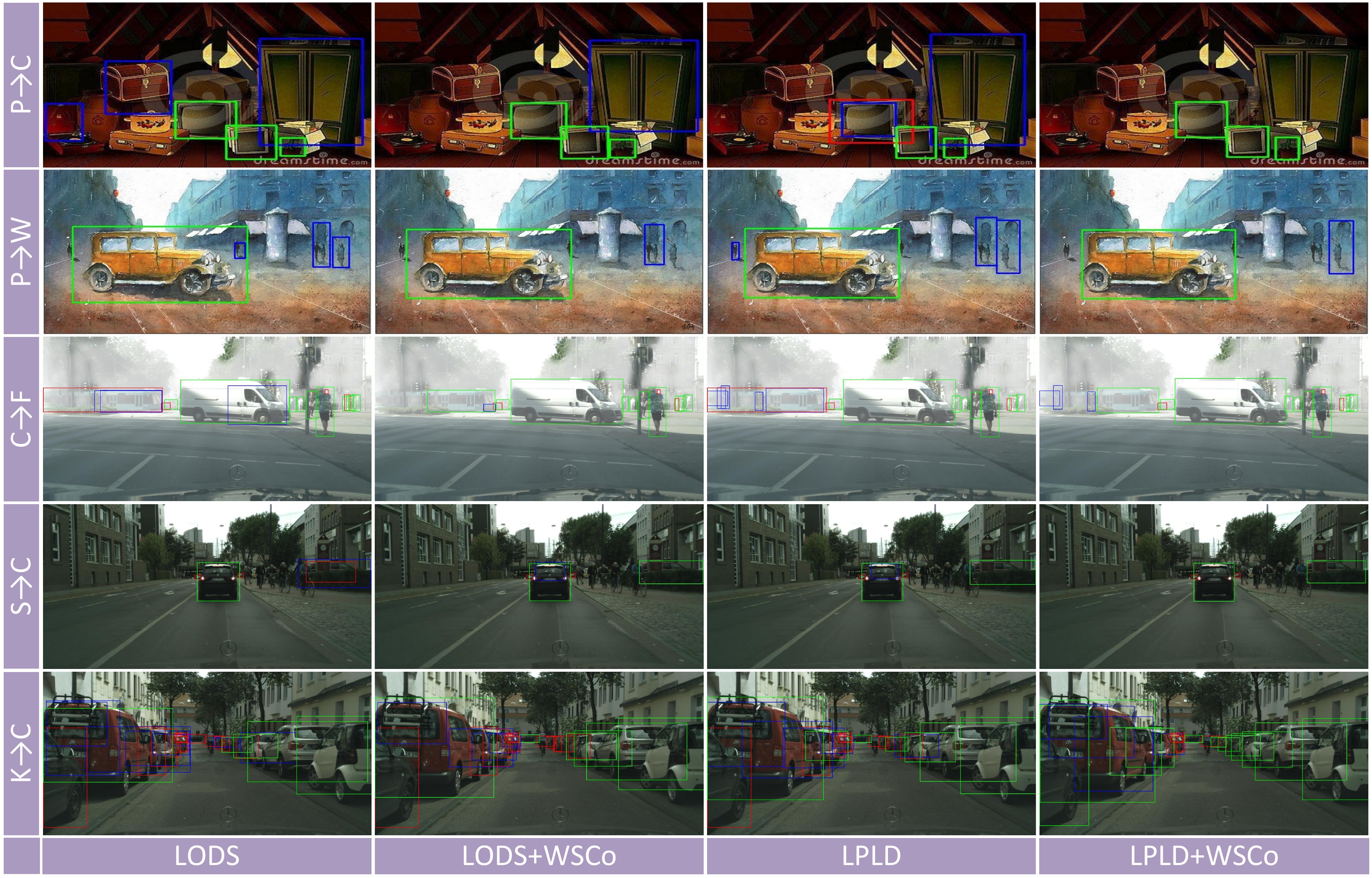}
\caption{Qualitative results on all five tasks: 
P $\rightarrow$ C ({Pascal $\rightarrow$ Clipart}), 
P $\rightarrow$ W ({Pascal $\rightarrow$ Watercolor}), 
C $\rightarrow$ F ({Cityscapes $\rightarrow$ FoggyCityscapes}), 
K $\rightarrow$ C ({KITTI $\rightarrow$ Cityscapes}), and 
S $\rightarrow$ C ({SIM10k $\rightarrow$ Cityscapes}).
\textcolor[RGB]{153,205,178}{\bf Top five rows:} Results of MT and IRG groups.
\textcolor[RGB]{171,154,192}{\bf Bottom five rows:} Results of LODS and LPLD groups. 
The \textcolor{green}{green}, \textcolor{red}{red}, and \textcolor{blue}{blue} bounding boxes represent true positives (TP), false negatives (FN), and false positives (FP), respectively. \textit{Zoom in for best view. }}
\label{fig:quanlity}
\end{figure*}

Formally, for the $M$ proposals of a weakly augmented image $\bar{\boldsymbol{I}}$, their weak instance features are $\bar{\boldsymbol{X}}=\{\bar{\boldsymbol{x}}_i\}_{i=1}^M$ and corresponding embeddings mapped by MNet are $\bar{\boldsymbol{Z}}=\{\bar{\boldsymbol{z}}_i\}_{i=1}^M$. 
For any weak instance feature $\bar{\boldsymbol{x}}_i$, we obtain vector $\boldsymbol{d}_{\bar{\boldsymbol{x}}_i} \in \mathbb{R}^C$ representing the distances from C clustering centers, conducting the adaptation-aware prototype-guided labeling upon $\bar{\boldsymbol{X}}=\{\bar{\boldsymbol{x}}_i\}_{i=1}^M$. 
Similarly, we obtain distance vector $\boldsymbol{d}_{\bar{\boldsymbol{z}}_i}\in \mathbb{R}^C$ by performing this method over $\bar{\boldsymbol{Z}}$. 
Thus, pseudo-category of the $i$-th instance in image $\bar{\boldsymbol{I}}$ is predicted by: 
\begin{equation}
\label{eqn:warmup}
\bar{y}_t = \arg\min_{j} (1 - \omega_t) \times \boldsymbol{d}_{\bar{\boldsymbol{x}}_i} + \omega_t \times \boldsymbol{d}_{\bar{\boldsymbol{z}}_i},~~j\in[0,~C-1], 
\end{equation}
where $\omega_t$ denotes the proportion weight of iteration $t$ in the first epoch. In practice, $\omega_t$ gradually increases from 0 to 1 with a step of $1/{T_1}$ where ${T_1}$ is the iteration number of the first epoch.

\section{Qualitative Comparison} \label{app-vis}
\textbf{Feature distribution.} 
For an intuitive analysis, we employed t-SNE tool to visualize the feature distribution based on the detection results of the SMT group on task Cityscapes $\rightarrow$ FoggyCityscapes.
Meanwhile, the source model (denoted as Source) and Oracle (trained on FoggyCityscapes with ground truth) are selected as comparisons.  
As illustrated in Fig.~\ref{fig:tsne-dis}, from Source to SMT+{\shortmodelname}, the aggregation becomes obvious gradually. 
Notably, SMT+{\shortmodelname} distribution shape closely resembles that of Oracle.

\textbf{Qualitative detection results.} 
To qualitatively verify the proposed {\shortmodelname} method, we visualize the typical detection results of the SMT, IRG, LODS, and LPLD comparison groups on all five transfer tasks. 
As shown in Fig.~\ref{fig:quanlity}, in all four comparison groups, the methods equipped with {\shortmodelname} are capable of detecting more objects while maintaining accuracy compared with these base methods. 
These results provide additional evidence that confirms the effectiveness of {\shortmodelname}.

\section{More Quantitative Results}
\label{app-quanti}

\textbf{Full results on task Pascal $\rightarrow$ Clipart.}
Tab.~\ref{tab:clipart50} is the supplement of average results on Pascal $\rightarrow$ Clipart (reported in Tab.~\ref{tab:water and clipar}), displaying the full detection results over the 20 categories. 
Specifically, in the ResNet-50 group, LPLD+{\shortmodelname} totally obtain best results on 6/20 categories, leading to the advantage on average accuracy.
On some cases, such as bottle, bus, bike and sofa, LODS+{\shortmodelname} has presents significant advantages over the previous methods.
When the backbone is switched to better ResNet-101, LODS+{\shortmodelname}'s advantage expands further, achieving best results on half categories.

\begin{table*}[t]
\caption{Full of results on task {Pascal $\rightarrow$ Clipart}.}
\renewcommand{\arraystretch}{1.3} 
    \centering
    \scriptsize
    \setlength{\tabcolsep}{0.45mm}{
    \begin{tabular}{clccccccccccccccccccccc|l}
        \toprule
        {} & Method & SF  & aero & bcle & bird & boat & bott & bus  & car  & cat  & chri & cow  & tabl & dog  & hors & bike & prsn & plnt & shep & sofa & trin & tv   & mAP \\

        \midrule
            \multirow{9}{*}{\rotatebox{90}{ResNet-101}} & Source~\cite{faster} & -- &26.8 & 28.8 & 23.4 & 24.1 & 41.9 & 31.4 & 28.5 & 4.9 & 32.0 & 11.0 & 29.8 & 4.3 & 41.2 & 53.3 & 43.7 & 42.0 & 13.1 & 19.9 & 37.1 & 34.4 & 28.6\\
            & SWDA~\cite{SWDA}  & \ding{55} & 26.2 & 48.5 & 32.6 & 33.7 & 38.5 & 54.3 & 37.1 & 18.6 & 34.8 & 58.3 & 17.0 & 12.5 & 33.8 & 65.5 & 61.6 & 52.0 & 9.3 & 24.9 & 54.1 & 49.1 & 38.1 \\
            & SAPNet~\cite{sapnet} & \ding{55} & 27.4 & 70.8 & 32.0 & 27.9 & 42.4 & 63.5 & 47.5 & 14.3 & 48.2 & 46.1 & 31.8 & 17.9 & 43.8 & 68.0 & 68.1 & 49.0 & 18.7 & 20.4 & 55.8 & 51.3 & 42.2 \\
            & PD~\cite{PD} & \ding{55} &41.5 &52.7& 34.5& 28.1& 43.7& 58.5& 41.8& 15.3& 40.1& 54.4& 26.7& 28.5& 37.7& 75.4& 63.7& 48.7& 16.5& 30.8& 54.5& 48.7& 42.1\\
            & UMT~\cite{UMT} & \ding{55} &39.6 & 59.1 &32.4 &35.0 &45.1 &61.9& 48.4 &7.5 &46.0 & 67.6 &21.4 & 29.5 &48.2 &75.9& 70.5 & 56.7 &25.9 &28.9& \textbf{39.4} &43.6 &44.1\\
            \cmidrule(lr){2-24}
            
            & SMT~\cite{meanteacher}  & \ding{51} & 33.3 & 43.4 & 23.9 & 35.7 & 48.9 & 65.3 & 36.4 & 6.1 & 41.0 & 18.9 & 26.7 & 13.1 & 37.4 & 60.4 & 44.2 & 41.2 & 24.5 & 15.2 & 57.0 & 43.3 & 35.8 \\

            &  \textbf{SMT+\shortmodelname} & \ding{51} & 42.2 & 61.3 & 30.0 & 35.9 & \textbf{51.7} & 59.6 & 46.9 & 9.1 & 44.2 & 17.8 & 20.4 & 17.3 & \textbf{48.7} & 83.4 & 59.9 & 46.1 & 22.2 & 21.9 & \textbf{62.7} & 44.6  & 41.3 {\tiny\color{blue} (\textuparrow 5.5)} \\
            
            & LODS~\cite{LODS}   & \ding{51} & \textbf{43.1}  & \textbf{61.4}& \textbf{40.1}& 36.8& 48.2& 45.8& 48.3& \textbf{20.4} & 44.8& 53.3& 32.5& \textbf{26.1}& 40.6& \textbf{86.3}& 68.5& 48.9& 25.4& 33.2& 44.0 & \textbf{56.5} &45.2 \\
            
            &  \textbf{LODS+\shortmodelname} & \ding{51} &46.1 & 54.6 & 32.0 & \textbf{39.4} & 47.1 & \textbf{78.7} & \textbf{49.4} & 3.0 & \textbf{54.0} & \textbf{58.7} & \textbf{43.7} & 24.4 & 47.0 & 82.5 & \textbf{68.6} & \textbf{49.4} & \textbf{27.3} & \textbf{39.9} & 51.9 & 48.9 & \textbf{47.3} {\tiny\color{blue} (\textuparrow 2.1)} \\
        \midrule
            \multirow{7}{*}{\rotatebox{90}{ResNet-50}} & Source~\cite{faster} & -- & 17.9 & 43.7 & 21.6 & 19.1 & 19.1 & 50.5 & 32.3 & 4.5 & 34.0 & 10.1 & \textbf{26.2} & 1.8 & 34.1 & 46.5 & 41.6 & 34.0 & 15.2 & 10.9 & 37.9 & 36.2 & 26.8 \\
            
            & SMT~\cite{meanteacher}  & \ding{51} & 21.6 & 56.3 & 24.6 & 17.5 & 28.0 & \textbf{76.1} & 36.7 & 9.1 & 32.0 & 11.0 & 23.2 & 10.3 & 34.3 & 62.1 & 39.2 & 43.7 & 9.1 & 16.9 & 39.3 & 36.3 & 31.4 \\

            &  \textbf{SMT+\shortmodelname} & \ding{51} & \textbf{23.6} & 54.8 & 27.1 & \textbf{26.4} & \textbf{47.1} & 59.9 & 39.9 & 9.1 & 37.5 & 16.3 & 24.3 & \textbf{17.2} & \textbf{41.0 }& 64.0 & \textbf{58.2} & \textbf{45.8} & 9.1 & 28.3 & 40.9 & 42.0 & 35.6 {\tiny\color{blue} (\textuparrow 4.2)} \\
            
            & IRG~\cite{IRG}  & \ding{51} & 20.3 & 47.3 &{27.3} & 19.7 & 30.5 & 54.2 & 36.2 & \textbf{20.6} & 35.1 & \textbf{20.6} & 20.2 & 12.3 & 28.7 & 53.1 & 47.5 & 42.4 & 9.1 & 21.1 & 42.3 & \textbf{50.3} & 31.5 \\
            
            & \textbf{IRG+\shortmodelname} & \ding{51} & 20.1 & 44.8 & \textbf{27.6} & 24.2 & 37.2 & 68.4 & 39.3 & 9.1 & 36.6 & 17.8 & 24.1 & 14.9 & 40.4 & 53.7 & 55.1 & 41.5 & 14.3 & 20.3 & 43.0 & 40.7 & 33.6 {\tiny\color{blue} (\textuparrow 2.1)}\\
            
            & LPLD~\cite{ECCV2024LPLD}  & \ding{51} & 18.9 & \textbf{66.1} & 25.6 & 21.1 & 37.6 & 61.7 & \textbf{45.4} & 9.1 & 33.7 & 11.2 & 20.5 & 14.5 & 32.3 &55.6 &57.0 &37.3 &18.2 &\textbf{31.7}& 39.5 & 42.6 &34.0 \\
            
            & \textbf{LPLD+\shortmodelname} & \ding{51} & 22.9 & 51.8 & 27.1 & 25.6 & 38.5 & 62.7 & 39.2 & 9.1 & \textbf{41.2} & 19.5 & 18.9 & 14.8 & 39.5 & \textbf{66.5} & 56.0 & \textbf{45.8} & \textbf{23.0} & 26.5 & \textbf{48.3} & 39.2 & \textbf{35.8} {\tiny\color{blue} (\textuparrow 1.8)} \\
        \bottomrule
    \end{tabular}}
    \label{tab:clipart50}
\end{table*}

\begin{figure*}[b]
\centering
\includegraphics[width=0.98\textwidth]{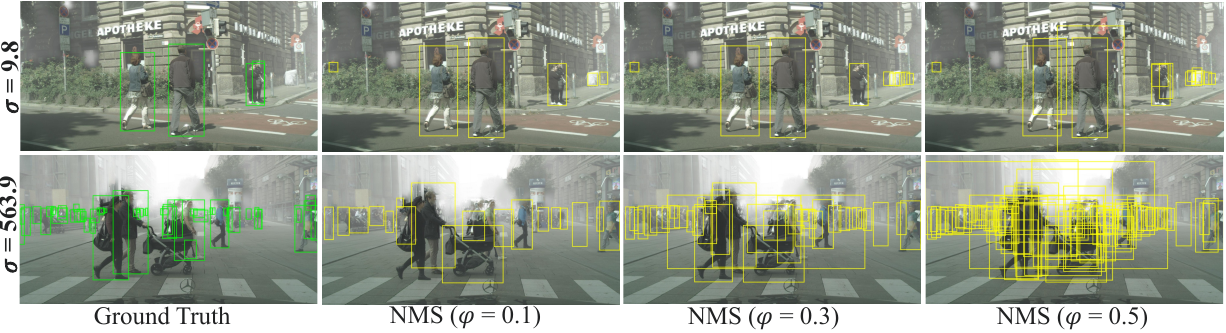}
\caption{Visualization of proposal-based image uncertainty estimation ($\sigma$) on task {Cityscapes $\rightarrow$ FoggyCityscapes}.
{\bf Top} and {\bf Bottom} provide the cases of $\sigma=9.8$ and $\sigma=563.9$, respectively.
}
\label{fig:vis-uncer}
\end{figure*}

 \section{Further Model Analysis}
\label{app-analy}

This section performs model analysis based on the SMT group, which includes SMT and SMT+{\shortmodelname}, without focusing on any other specific designs.
This approach allows our evaluation to emphasize the proposed {\shortmodelname} method.




\textbf{Analysis of proposal-based uncertainty estimation.}
For an intuitive understanding of the working mechanism of the image uncertainty estimation, this part visualizes the image uncertainty estimation of two typical images in Fig.~\ref{fig:vis-uncer}.  
As shown in the top row, the number of proposals has increased slightly since the light foggy case is similar to the source domain, which has a clear view.  
In contrast, the number of proposals increases significantly when the unfamiliar situation involves heavy fog (see the bottom row). 
These observations are consistent with our expectations, and they are also reflected in the estimated values.

On the other hand, the visualization results provide insight into the effectiveness of image uncertainty estimation. 
In the low-uncertainty case ($\sigma=9.8$), the background proposals are clearly distinguished from the target proposals, which can offer important visual clues for differentiation. Therefore, it makes sense to include these background proposals in the contrastive method. In contrast, the high-uncertainty case ($\sigma=563.9$) shows tightly clustered proposals, which may introduce a significant amount of noise. For instance, proposals crossing pedestrians are treated as background. If these proposals are incorporated into the contrastive method, they could lead to a misunderstanding of the pedestrian conception.

\begin{figure}[t]
    \centering
    \makeatletter\def\@captype{figure}
    \renewcommand\tabcolsep{2.5pt}
    \renewcommand\arraystretch{0.95}
    \centering
    \begin{center}
        \includegraphics[width=0.95\linewidth]{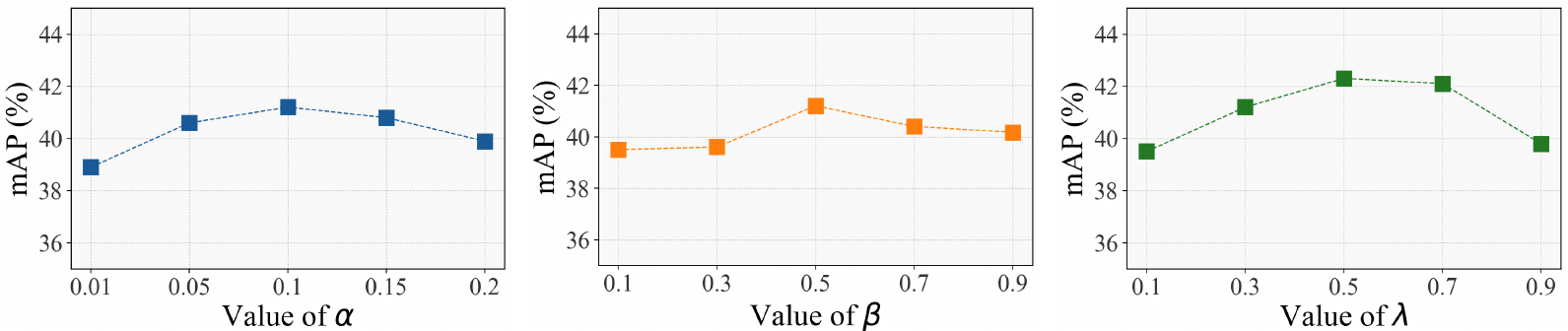}
    \end{center}
    \caption{
    The mAP varying curve of three main hyper-parameters in {\shortmodelname} based on the task Pascal $\rightarrow$ Clipart. 
    {\bf Left}, {\bf Middle} and {\bf Right} are results of the parameter $\alpha$, $\beta$, and $\lambda$, respectively.
    }
    \label{tab:params}
\end{figure}


\textbf{Analysis of hyper-parameters sensitivity.} 
In this part, we discuss the impact of the three hyperparameters, including trading-off parameter $\alpha$, $\beta$ and $\lambda$ in Eq.~\eqref{eqn:loss-seman-cali}, Eq.~\eqref{eqn:obj-final}, and Eq.~\eqref{eqn:loss-cis}.  
All results are obtained based on the adaptation task Cityscapes $\rightarrow$ FoggyCityscapes.  
As shown in Left of Fig.~\ref{tab:params}, our model can maintain relatively stable results over a wide range of $\alpha$ (0.05$\sim$0.15). 
The same phenomenon is observed in Middle and Right of Fig.~\ref{tab:params}. 
As $\beta \in \left[0.3, 0.7\right]$ and $\lambda \in \left[0.3, 0.7\right]$, there are not significant performance drop in mAP. 
These results indicate that our {\shortmodelname} is insensitive to the parameters.  

\begin{wrapfigure}{r}{0.6\textwidth}
    \centering
    \makeatletter\def\@captype{figure}
    \renewcommand\tabcolsep{3.5pt}
    \renewcommand\arraystretch{0.95}
    \centering
    \begin{center}
        \includegraphics[width=0.47\linewidth]{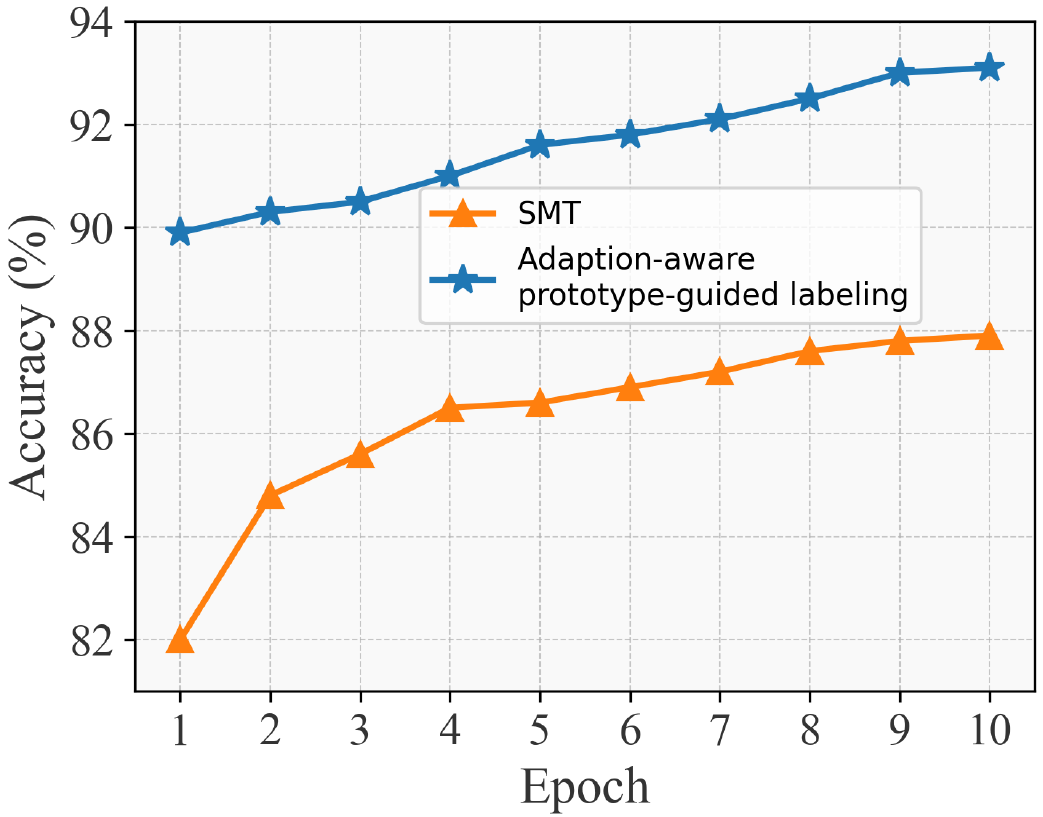}
        \includegraphics[width=0.47\linewidth]{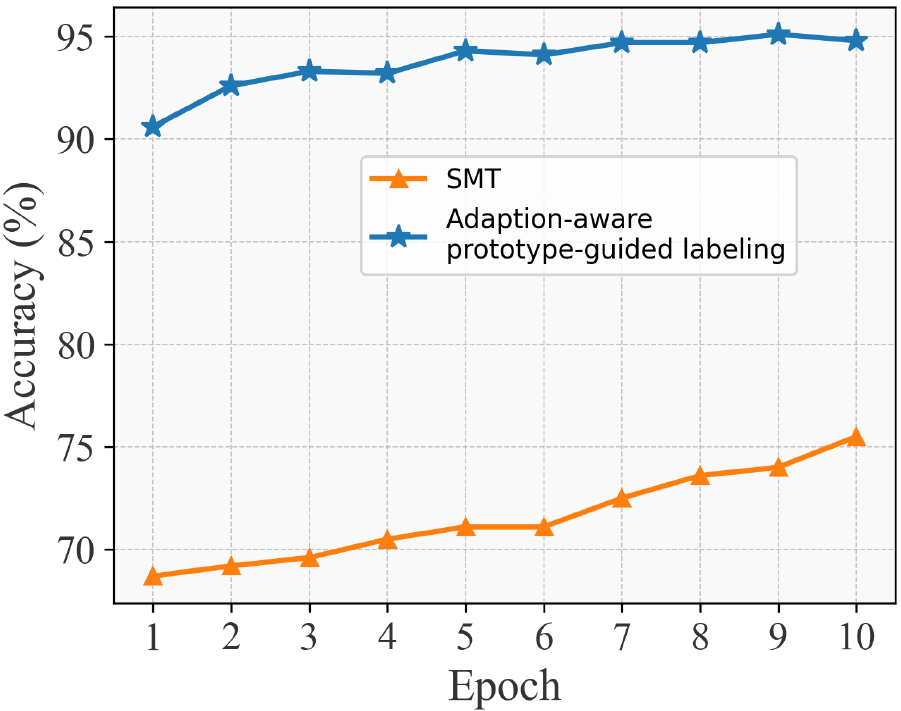}
    \end{center}
    \vspace{-0.2em}
    \caption{Classification accuracy of foreground instances on transfer task Cityscapes $\rightarrow$ FoggyCityscapes ({\bf Left}), Pascal $\rightarrow$ Watercolor ({\bf Right}).}
    \label{tab:cluster}
\end{wrapfigure}
\textbf{Analysis of adaptation-aware prototype guided labeling.}
To evaluate the effectiveness of the adaptation-aware prototype-guided labeling, we conducted a quantitative analysis on tasks Cityscapes $\rightarrow$ FoggyCityscapes and Pascal $\rightarrow$ Watercolor, as presented in Fig.~\ref{tab:cluster}. 
Specifically, we compared the classification accuracy of foreground instances in proposals (with an IoU greater than 0.5 against the ground truth) across the two tasks.
Meanwhile, the accuracy predicted by SMT is taken as a comparison. 
It is seen that the prototype-guided labeling significantly outperforms SMT, indicating that our method can effectively extract valuable information from the weak instance embeddings.

\begin{table}[t]
    \begin{minipage}[t]{0.55\textwidth}
        \centering
        \caption{
        Error bars on task Pascal$\rightarrow$Watercolor. SD is short for Standard Deviation.}
        
        \renewcommand\tabcolsep{1pt}
        \renewcommand\arraystretch{1.2}
        \scriptsize
        \begin{tabular}{l|ccccc|cc}
        \toprule
        & Default & \multicolumn{4}{c}{Randomly selected} \vline\\
        \cmidrule(lr){2-2} \cmidrule(lr){3-6}
        Method/Seed        &17731508  &18547981  &64352569 &14378526  &15498567  &Mean  &SD \\
        \midrule
        SMT+{\shortmodelname} &56.9  &57.1  &56.6  &56.8  &56.7  &56.82  &0.192 \\
        \bottomrule
        \end{tabular}
        \label{tab:seed}
    \end{minipage}
    ~~~~~
    \begin{minipage}[t]{0.40\textwidth}
        \centering
        \caption{
        {Training resource demands on task Cityscapes $\rightarrow$ Foggycityscapes.}
        }
        \renewcommand\tabcolsep{3.5pt}
        \renewcommand\arraystretch{1.4} 
        \scriptsize
        \begin{tabular}{l|cc}
            \toprule
            {Item / Method}         & SMT~\cite{meanteacher}  & SMT+{\shortmodelname} \\
            \midrule
            {GPU memory cost (GB)}  & {5.23}  & {6.81} {\tiny\color{blue} (\textuparrow 1.58)} \\
            {Training times (s)}    & {0.19}  & {0.38} {\tiny\color{blue} (\textuparrow 0.19)} \\
            \bottomrule
        \end{tabular}
        \label{tab:trn-res}
    \end{minipage}
\end{table}

\textbf{Error bar analysis.}
To ensure the reproduction of experimental results, we use a fixed randomly generated random seed 17731508 in all experiments. 
Here, in order to analyze the error bars introduced by this, SMT+{\shortmodelname} is run under four random seeds that we randomly pick up.
With the five seeds, we calculate the average and standard deviation of their mAP results.
As shown in Tab.~\ref{tab:seed}, the seed variation leads to a tiny performance vibration (0.192), indicating our method is insensitive to the selection of random seeds.

\textbf{Training resource demands.} 
For a plug-in, the training resource demands are an important deployment-related feature. 
In this part, we conduct training resource comparison on transfer task Cityscapes $\rightarrow$ FoggyCityscapes. 
The results shown in Tab.~\ref{tab:trn-res} indicate that our {\shortmodelname} does not incur significant additional training costs and requires a similar amount of computational resources as the base method.

\section{Limitation}
In this paper, we discuss the artificial inter-category fusion problem in the context of two-stage detection architectures. 
Conceptually, it is applicable to one-stage architectures, but there is an extra need to generate intermediate proposals. This will increase the overall change. We will consider how to expand in this direction as future work.

\clearpage

\end{document}